\documentclass[10pt,twocolumn,letterpaper]{article}

\usepackage[pagenumbers]{wacv} 

\usepackage{algorithm}
\usepackage{multirow}
\usepackage{comment}
\usepackage{float}
\usepackage{wasysym}
\usepackage{cuted}
\usepackage{pgfplots}
\usepackage{tikz}
\usetikzlibrary{shapes.geometric, positioning, decorations.pathreplacing, calc, intersections, pgfplots.groupplots}
\usepackage[noend]{algpseudocode}
\usepackage{colortbl}
\usepackage{adjustbox}
\usepackage{fancyhdr}

\definecolor{wacvblue}{rgb}{0.21,0.49,0.74}
\usepackage[hidelinks, pagebackref,breaklinks,colorlinks,allcolors=wacvblue]{hyperref}
\usepackage{url}

\def\confName{WACV}
\def\confYear{2026}

\title{Restora-Flow: Mask-Guided Image Restoration with Flow Matching}

\author{Arnela Hadzic$^{1}$, Franz Thaler$^{1,2,3}$, Lea Bogensperger$^{4}$, Simon Johannes Joham$^{1}$, Martin Urschler$^{1}$\vspace{0.2cm}\\
{\small $^{1}$Institute for Medical Informatics, Statistics and Documentation, Medical University of Graz, Graz, Austria}\\[-2pt]
{\small $^{2}$Division of Medical Physics and Biophysics, Medical University of Graz, Graz, Austria}\\[-2pt]
{\small $^{3}$Institute of Visual Computing, Graz University of Technology, Graz, Austria}\\[-2pt]
{\small $^{4}$Department of Quantitative Biomedicine, University of Zurich, Zurich, Switzerland}
}

\pgfplotsset{compat=1.18}
\begin{document}

\fancypagestyle{plain}{
  \fancyhf{} 
  \fancyhead[L]{Accepted for IEEE/CVF Winter Conference on Applications of Computer Vision (WACV) 2026}
  \fancyfoot[C]{\thepage} 
}

\maketitle

\begin{strip}
\begin{minipage}{\textwidth}
\vspace{-0.5cm}
\adjustbox{center=16cm}{
    \scalebox{0.89}{\definecolor{repaint}{HTML}{955196}
\definecolor{otode}{HTML}{87A96B}
\definecolor{flowpriors}{HTML}{CFB53B}
\definecolor{dflow}{HTML}{FF9966}
\definecolor{pnpflow}{HTML}{6699CC}
\definecolor{restoraflow}{HTML}{CC3333}

\definecolor{denoising}{HTML}{9466FF}
\definecolor{boxinpainting}{HTML}{69C7F8}
\definecolor{superresolution}{HTML}{D93F50}
\definecolor{randominpainting}{HTML}{FF9966}

\newcommand{\repaintmark}{asterisk}
\newcommand{\ddnmmark}{+}
\newcommand{\otodemark}{*}
\newcommand{\flowpriorsmark}{square*}
\newcommand{\dflowmark}{diamond*}
\newcommand{\pnpflowmark}{triangle*}
\newcommand{\restoraflowmark}{pentagon*}

\begin{tikzpicture}
    \node[scale=0.9] (plot) at (0, 0) {
        \begin{tikzpicture}
            \begin{groupplot}[
                group style={
                    group name=group,
                    group size=2 by 1,
                    horizontal sep=10pt
                },
                ylabel={\small Score},
                ymin=0, ymax=0.5,
                yticklabel style={/pgf/number format/fixed, /pgf/number format/precision=5},
                grid=major,
            ]
        
            \nextgroupplot[xmax=50,
                xtick={0, 10, 20, 30, 40, 50},
                xticklabels={0, 10, 20, 30, 40},
                ytick={0, 0.1, 0.2, 0.3, 0.4, 0.5},
                axis y line*=left,
                width=6cm,
                scale only axis,
                height=5cm,
                mark options={scale=1.5},
            ]
    
            \addplot[only marks, mark=\ddnmmark, mark size=1.75, color=denoising] coordinates {(13.74, 0.170)};

            \addplot[only marks, mark=\otodemark, mark size=1.75, color=denoising] coordinates {(5.54, 0.078)};
            \addplot[only marks, mark=\flowpriorsmark, mark size=1.5, color=denoising] coordinates {(67.10, 0.153)};
            \addplot[only marks, mark=\dflowmark, color=denoising] coordinates {(44.45, 0.184)};
            \addplot[only marks, mark=\pnpflowmark, color=denoising] coordinates {(9.86, 0.165)};
            \addplot[only marks, mark=\restoraflowmark, mark options={draw=black, line width=0.05mm, fill=denoising}, mark size=4] coordinates {(0.72, 0.051)};
        
            \addplot[only marks, mark=\repaintmark, mark size=1.75, color=boxinpainting] coordinates {(86.23, 0.043)};
            \addplot[only marks, mark=\ddnmmark, mark size=1.75, color=boxinpainting] coordinates {(13.74, 0.048)};
            
            \addplot[only marks, mark=\otodemark, mark size=1.75, color=boxinpainting] coordinates {(6.94, 0.048)};
            \addplot[only marks, mark=\flowpriorsmark, mark size=1.5, color=boxinpainting] coordinates {(67.05, 0.054)};
            \addplot[only marks, mark=\dflowmark, color=boxinpainting] coordinates {(126.09, 0.112)};
            \addplot[only marks, mark=\pnpflowmark, color=boxinpainting] coordinates {(9.86, 0.124)};
            \addplot[only marks, mark=\restoraflowmark, mark options={draw=black, line width=0.05mm, fill=boxinpainting}, mark size=4] coordinates {(3.96, 0.047)};
        
            \addplot[only marks, mark=\repaintmark, mark size=1.75, color=superresolution] coordinates {(86.23, 0.139)};
            \addplot[only marks, mark=\ddnmmark, mark size=1.75, color=superresolution] coordinates {(13.74, 0.462)};
            
            \addplot[only marks, mark=\otodemark, mark size=1.75, color=superresolution] coordinates {(7.28, 0.285)};
            \addplot[only marks, mark=\flowpriorsmark, mark size=1.5, color=superresolution] coordinates {(67.30, 0.271)};
            \addplot[only marks, mark=\dflowmark, color=superresolution] coordinates {(261.84, 0.123)};
            \addplot[only marks, mark=\pnpflowmark, color=superresolution] coordinates {(46.26, 0.180)};
            \addplot[only marks, mark=\restoraflowmark, mark options={draw=black, line width=0.05mm, fill=superresolution}, mark size=4] coordinates {(14.48, 0.158)};
        
            \addplot[only marks, mark=\repaintmark, mark size=1.75, color=randominpainting] coordinates {(86.23, 0.034)};
            \addplot[only marks, mark=\ddnmmark, mark size=1.75, color=randominpainting] coordinates {(13.74, 0.065)};
            \addplot[only marks, mark=\otodemark, mark size=1.75, color=randominpainting] coordinates {(7.28, 0.094)};
            \addplot[only marks, mark=\flowpriorsmark, mark size=1.5, color=randominpainting] coordinates {(67.69, 0.046)};
            \addplot[only marks, mark=\dflowmark, color=randominpainting] coordinates {(266.18, 0.056)};
            \addplot[only marks, mark=\pnpflowmark, color=randominpainting] coordinates {(19.15, 0.042)};
            \addplot[only marks, mark=\restoraflowmark, mark options={draw=black, line width=0.05mm, fill=randominpainting}, mark size=4] coordinates {(7.48, 0.034)};





            \node[scale=0.7, inner sep=2pt, draw=black, fill=white, anchor=north west, text width=3.2cm] at (axis cs:28.3, 0.495) {%
              Methods: \\
              \vspace{1mm}
              \begin{tabular}{c @{\hspace{2pt}} l }

                \scalebox{1.2}{$\ast$} & RePaint \cite{lugmayr2022repaint} \\ \scalebox{0.8}{$+$} & DDNM\textsuperscript{+} \cite{wang2023zeroshot} \\ 

                \scalebox{1.2}{$\circ$} & OT-ODE \cite{pokle2024training} \\ \scalebox{0.8}{$\square$} & Flow-Priors \cite{zhang2024flow} \\
                \scalebox{1.2}{$\diamond$} & D-Flow \cite{benhamu2024dflow} \\ \scalebox{0.8}{$\triangle$} & PnP-Flow \cite{martin2024pnp}  \\
                \scalebox{1.2}{$\pentagon$} & \textbf{Restora-Flow} \\
              \end{tabular}
            };

            \nextgroupplot[xmin=50,xmax=300,
                xtick={50, 100, 200, 300},
                clip=false,
                ylabel={},
                ytick={},
                yticklabels={},
                axis y line*=right,
                width=2.75cm,
                scale only axis,
                height=5cm,
                mark options={scale=1.5}
            ]


            \addplot[only marks, mark=\flowpriorsmark, mark size=1.5, color=denoising] coordinates {(67.10, 0.153)};
        
            \addplot[only marks, mark=\repaintmark, mark size=1.75, color=boxinpainting] coordinates {(86.23, 0.043)};
            
            \addplot[only marks, mark=\flowpriorsmark, mark size=1.5, color=boxinpainting] coordinates {(67.05, 0.054)};
            \addplot[only marks, mark=\dflowmark, color=boxinpainting] coordinates {(126.09, 0.112)};
        
            \addplot[only marks, mark=\repaintmark, mark size=1.75, color=superresolution] coordinates {(86.23, 0.139)};
            
            \addplot[only marks, mark=\flowpriorsmark, mark size=1.5, color=superresolution] coordinates {(67.30, 0.271)};
            \addplot[only marks, mark=\dflowmark, color=superresolution] coordinates {(261.84, 0.123)};
        
            \addplot[only marks, mark=\repaintmark, mark size=1.75, color=randominpainting] coordinates {(86.23, 0.034)};
            \addplot[only marks, mark=\flowpriorsmark, mark size=1.5, color=randominpainting] coordinates {(67.69, 0.046)};
            \addplot[only marks, mark=\dflowmark, color=randominpainting] coordinates {(266.18, 0.056)};

            \node[scale=0.7, inner sep=2pt, draw=black, fill=white, anchor=north west, text width=3.2cm] at (axis cs:82, 0.495) {%
              Tasks:\\
              \vspace{1mm}
              \begin{tabular}{l}
                {\color{denoising} Denoising} \\ 
                {\color{superresolution} Super-resolution} \\
                {\color{boxinpainting} Box inpainting} \\ 
                {\color{randominpainting} Random inpainting} \\
              \end{tabular}
            };

            \node[inner sep=0, anchor=north west] (cat_denoising) at (axis cs:340, 0.5)
            {\includegraphics[height=5cm]{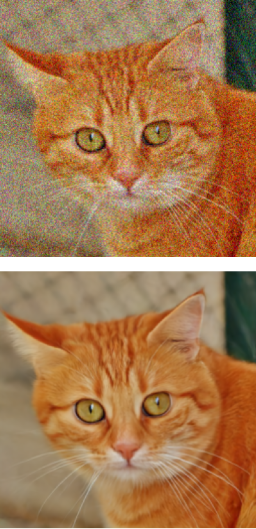}};

            \node[inner sep=0, anchor=north west] (celeba_superresolution) at (axis cs:574, 0.5)
            {\includegraphics[height=5cm]{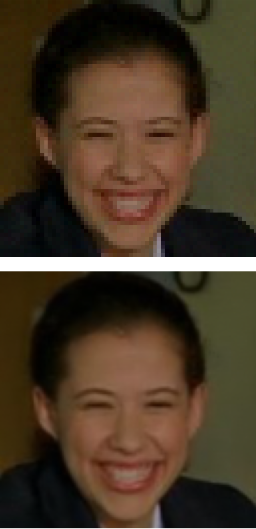}};

            \node[inner sep=0, anchor=north west] (coco_random_inpainting) at (axis cs:808, 0.5)
            {\includegraphics[height=5cm]{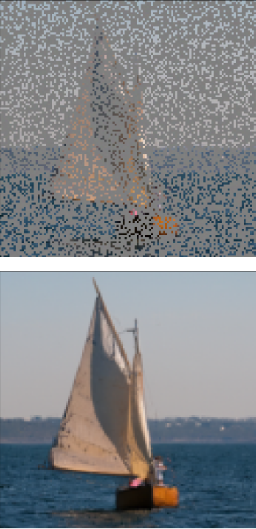}};

            \node[inner sep=0, anchor=north west] (medical_box_inpainting) at (axis cs:1042, 0.5)
            {\includegraphics[height=5cm]{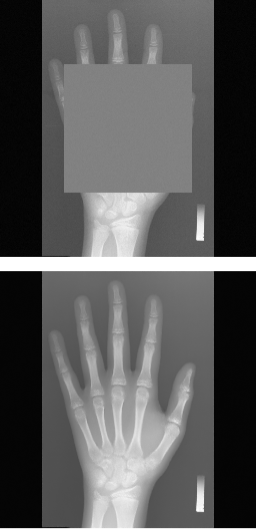}};

            \end{groupplot}

            \node (title) at ($(group c1r1.north)!0.5!(group c2r1.north)+(-0.8,0.4cm)$) {\small AFHQ-Cat: LPIPS $\downarrow$};
            \node (xlabel) at ($(group c1r1.south)!0.5!(group c2r1.south)+(-0.8,-0.7cm)$) {\small Time (s)};

            \node[inner sep=0, anchor=north west] (tilde) at ($(group c1r1.east)+(0.02cm,0)$) {$\sim$};

            \node[inner sep=0] (denoising) at ($(cat_denoising |- xlabel)$) {\small Denoising};
            \node[inner sep=0] (superresolution) at ($(celeba_superresolution |- xlabel)$) {\small Super-resolution};
            \node[inner sep=0] (randominpainting) at ($(coco_random_inpainting |- xlabel)$) {\small Random inpainting};
            \node[inner sep=0] (randominpainting) at ($(medical_box_inpainting |- xlabel)$) {\small Box inpainting};
        
        \end{tikzpicture}
    };

\end{tikzpicture}
}
}

\captionof{figure}{
We propose \textbf{Restora-Flow}, a mask-guided image \textbf{restora}tion method based on \textbf{flow} matching with an integrated trajectory correction mechanism.
Restora-Flow offers excellent performance in both reconstruction quality and processing efficiency when compared to state-of-the-art methods for various tasks.
This is demonstrated in the results showing LPIPS perceptual quality score vs. processing time for the AFHQ-Cat dataset (left).
Exemplary qualitative reconstruction results of Restora-Flow are shown across four different tasks and evaluated on four datasets with distinct characteristics (right).}
\label{fig:overview}
\end{minipage}
\end{strip}

\begin{abstract}
Flow matching has emerged as a promising generative approach that addresses the lengthy sampling times associated with state-of-the-art diffusion models and enables a more flexible trajectory design, while maintaining high-quality image generation. This capability makes it suitable as a generative prior for image restoration tasks.
Although current methods leveraging flow models have shown promising results in restoration, some still suffer from long processing times or produce over-smoothed results.
To address these challenges, we introduce Restora-Flow, 
a training-free method that guides flow matching sampling by a degradation mask and incorporates a trajectory correction mechanism to enforce consistency with degraded inputs. 
We evaluate our approach on both natural and medical datasets across several image restoration tasks involving a mask-based degradation, i.e., inpainting, super-resolution and denoising. We show superior perceptual quality and processing time compared to diffusion and flow matching-based reference methods. 
Code is available at {\small \url{https://github.com/imigraz/Restora-Flow}}.
\end{abstract}    
\section{Introduction}
\label{sec:intro}
Many important image restoration tasks, such as image denoising, super-resolution and inpainting, can be framed as inverse problems involving mask-based degradation. The objective is
to recover an original image $x \in \mathbb{R}^d$ from its degraded observation $z = Hx + \xi$, where $H$ represents a degradation operator using a mask $m$, and $\xi \sim \mathcal{N}(0,\sigma^2 I)$ denotes measurement noise.
The goal of image restoration is twofold: the restored image should exhibit high visual quality while also maintaining fidelity to the input data, ensuring consistency with the degraded observation.

Over the last few years, diffusion models have become a standard approach for generative modeling due to their ability to learn complex data distributions and generate high-quality images~\cite{sohl2015deep,ho2020denoising,song2021score,dhariwal2021diffusion,rombach2022high}. This has led to a growing interest in utilizing these generative models as unsupervised priors in inverse problems~\cite{daras2024survey}, with numerous successful applications to guide the restoration process~\cite{kawar2021snips,kawar2022denoising,lugmayr2022repaint,chung2023diffusion,song2023pseudoinverse,wang2023zeroshot,zhu2023denoising,mardani2024a}. 
However, a significant limitation of these methods is the long sampling time due to highly curved sampling trajectories~\cite{bartosh2025neural}, as well as the inherent challenges arising from the intermediate noisy steps.

More recently, Flow Matching (FM) has emerged as an alternative approach in generative modeling~\cite{lipman2022flow,liu2023flow}, characterized by its ability to maintain significantly straighter trajectories during data generation compared to diffusion models. FM defines the generative process using conditional optimal transport paths through Ordinary Differential Equations (ODEs). Recent research shows that FM achieves faster training and sampling while still generating high-quality images~\cite{esser2024scaling,zhang2024mutli,bogensperger2024flowsdf}, making it a promising direction for addressing image restoration tasks. To date, only a few studies have investigated the use of flow-based priors in solving image restoration~\cite{pokle2024training,zhang2024flow,benhamu2024dflow,martin2024pnp}. While these studies show promising results, some still suffer from relatively long processing times~\cite{zhang2024flow,benhamu2024dflow}, others yield over-smoothed samples~\cite{martin2024pnp}, or tend to produce artifacts~\cite{pokle2024training}.

In this work, we introduce Restora-Flow, a training-free, flow matching-based approach, specifically designed to address mask-based image restoration tasks.
Our approach conditions the sampling process of an unconditional flow matching model through mask-guided fusion, ensuring that regions unaffected by the degradation operator $H$ are preserved during sampling. The samples are then refined using a trajectory correction mechanism to enforce consistency with degraded observations.
When evaluated on several image restoration tasks using various datasets, Restora-Flow achieves the new state-of-the-art when simultaneously assessing both reconstruction quality and processing time.
Our contributions are summarized as follows:
\begin{itemize}
    \item We introduce Restora-Flow, a training-free algorithm for addressing mask-based inverse problems, utilizing unconditional flow prior models and mask-guided fusion for flow matching sampling.
    \item To enhance fidelity of the restoration process, we introduce a novel correction mechanism that guides the flow trajectory toward better alignment with observed data.
    \item We perform a comprehensive evaluation using computer vision and medical datasets across various settings for inpainting, super-resolution and denoising tasks. The results show that our method achieves superior perceptual quality compared to  related approaches~\cite{pokle2024training,zhang2024flow,benhamu2024dflow,martin2024pnp}, while also demonstrating the fastest processing time.
\end{itemize}

\section{Related Work}
\label{sec:related_work}

With the advent of deep learning, deep neural networks have become widely utilized in image restoration tasks. Various methods have been proposed to learn a direct mapping from degraded measurements to clean images by minimizing a reconstruction loss~\cite{dong2015image,zhang2017learning,tao2018scale,zamir2021multi}. However, these approaches typically require datasets containing paired degraded and clean images and necessitate retraining for each new task. To overcome these limitations, recent research has investigated the use of deep generative priors, such as diffusion or flow matching-based priors, due to their ability to model complex data statistics~\cite{jalal2021robust,song2021solving}. These generative approaches do not rely on paired datasets, enabling their application across various different image restoration tasks using the same prior knowledge.

Several diffusion-based approaches for image restoration have introduced guidance strategies to ensure consistency with degraded observations during the generation process. DDRM~\cite{kawar2022denoising} 
tackles linear inverse problems by employing singular value decomposition of the degradation operator~\( H \). Similarly, DDNM\textsuperscript{+}~\cite{wang2023zeroshot} addresses inverse problems in a zero-shot manner by utilizing range-null space decomposition as the guidance function, whereas 
RePaint~\cite{lugmayr2022repaint} uses unmasked regions to guide the diffusion process for inpainting tasks. To ensure consistency between denoising results and degraded measurements, \(\Pi\)GDM~\cite{song2023pseudoinverse} incorporates a vector-Jacobian product as additional guidance. In contrast, RED-Diff~\cite{mardani2024a} formulates image restoration as an optimization problem, 
minimizing a measurement consistency loss while applying score-matching regularization.

More recently, the application of flow matching as a prior in image restoration has gained considerable attention. OT-ODE~\cite{pokle2024training} incorporates the gradient correction term from \(\Pi\)GDM to guide the flow-based generation process. Their findings demonstrate that images restored using flow matching priors have superior perceptual quality compared to those generated via diffusion paths, as well as those produced by RED-Diff or \(\Pi\)GDM. Similarly to RED-Diff, D-Flow~\cite{benhamu2024dflow} formulates the image restoration process as a source point optimization problem. The objective is to minimize the cost function associated with the initial point in the flow matching framework. However, this optimization requires backpropagation through an ODE solver, resulting in relatively long sampling times (\ie, 5 to 15 minutes per sample, as reported in~\cite{benhamu2024dflow}). In contrast, Flow-Priors~\cite{zhang2024flow} decomposes the flow's trajectory into several local objectives and employs Tweedie's formula~\cite{efron2011tweedie} to sequentially optimize these objectives through gradient steps, resulting in reduced sampling time compared to D-Flow. Instead, PnP-Flow~\cite{martin2024pnp} combines Plug-and-Play methods~\cite{venkatakrishnan2013plug} with flow matching without requiring backpropagation, however, it tends to produce over-smoothed results.

\section{Method}\label{sec:method}
We introduce Restora-Flow, a novel training-free method for mask-based image restoration that leverages flow matching generative priors. At its core, Restora-Flow performs time-dependent ODE-based sampling guided by a mask and employs a trajectory correction mechanism that refines the generative path toward the data manifold. This correction is designed to align the evolving sample with the degraded input, thereby improving consistency between restored and observed known regions. Empirically, we find that a single correction step per ODE iteration is sufficient to ensure this consistency, enabling both high reconstruction quality and fast processing times (see~\cref{fig:overview,fig:qualitative_correction_steps}).

\subsection{Flow Matching (FM)}\label{sec:fm}
In flow matching~\cite{lipman2022flow,liu2023flow}, the idea is to learn a velocity field \(v_{\theta,t}\) of the probability flow \( \Psi_t \).
This field governs the transformation of a simple base distribution at \( t=0 \) into the target distribution \( \mathrm{p}(x) \) at \( t=1 \) by defining a continuous-time trajectory that maps between them.
For simulation-free training, the conditional FM loss~\cite{lipman2022flow} is used:
\begin{equation}
    \min_\theta \mathbb{E}_{t,x_1,x_0}\Bigl[\frac{1}{2}\Bigl\Vert v_{\theta,t}(\Psi_t(x_0)) - (x_1 - x_0)\Bigr\Vert^2\Bigr],
\end{equation}
with \( t \sim \mathcal{U}[0,1] \), \( x_1 \sim \mathrm{p}(x) \), \( x_0 \sim \mathcal{N}(0,I) \), and the conditional flow \( \Psi_t(x_0) = (1-t)x_0 + t\,x_1 \). The learned vector field \( v_{\theta,t} \) models the distribution's change in time, from which we can sample by integrating the corresponding ODE:
\begin{equation}
   \frac{\mathrm{d}}{\mathrm{d}t} \Psi_t(x) = v_{\theta,t}(\Psi_t(x)). 
\end{equation}
For example, using the explicit Euler integration scheme with \( t = 0, \dots, 1-\Delta_t \), the estimate is updated via
\begin{equation} \label{eq:fm_sampling}
    x_{t+\Delta_t} = x_t + \Delta_t \, v_{\theta,t}(x_t).
\end{equation}

\subsection{Restora-Flow Algorithm}\label{sec:restora-flow}
The restoration of an unknown image $x$ from its degraded observation $z$ can be formulated as a maximum a posteriori (MAP) estimation problem:
\begin{equation} \label{eq:map}
\hat{x} = \arg\min_{x} \mathcal{D}(H x, z) + \mathcal{R}_\theta(x), 
\end{equation}
where $H$ is the degradation operator, $\mathcal{D}(Hx,z)$ is a data fidelity term, and $R_\theta(x)$ is a prior encoding learned prior knowledge about images, parameterized by $\theta$.

When employing flow matching as the prior  \( \mathcal{R}_\theta(x) \), the generation of samples via only \cref{eq:fm_sampling} does not yield a minimizer of \cref{eq:map}. 
Therefore, the sampling must be guided toward the MAP solution by incorporating the degraded observation \( z \) into the sampling process. To optimize \cref{eq:map}, one can use either explicit gradient steps on the data term $\nabla_x \mathcal{D}(Hx,z)$ or implicit strategies, such as mask-guided update used in inpainting \cite{lugmayr2022repaint}. We opt for mask-guidance, due to its inherent relationship with inpainting and other image restoration tasks involving mask-based degradations.

\begin{figure}[t]
    \centering
    \includegraphics[width=0.99\linewidth]{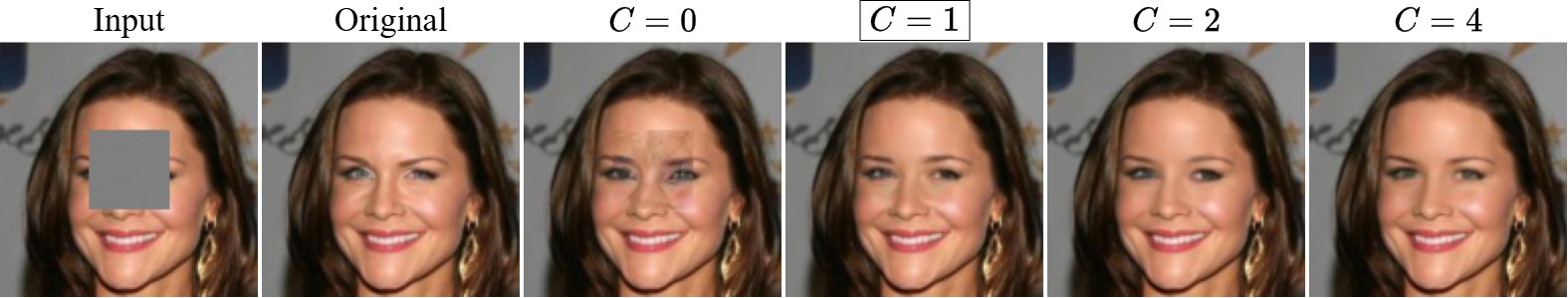}
    \caption{Restora-Flow samples with and without correction steps. Empirically, one correction step ($C=1$) offers the best trade-off between high reconstruction quality and fast processing.}
    \label{fig:qualitative_correction_steps}
\end{figure}

\paragraph{Incorporating mask-guidance} 
Mask-guidance, introduced by RePaint \cite{lugmayr2022repaint}, fuses the time-dependent variable $x_t$ with the unmasked portion of the original image $z$. First, $z$ is adapted to match the noise level contained in the flow estimate $x_t$ through the convex combination 
\begin{equation}\label{eq:noisy_z}
    z' = t z + (1-t) \epsilon, \quad  \epsilon \sim \mathcal{N}(0,I) ,
\end{equation}
and is subsequently fused with $x_t$ via
\begin{equation}\label{eq:fusion}
     x_{t}' = m \odot z' + (1-m) \odot x_{t} ,
\end{equation}
where $\odot$ denotes element-wise multiplication and $m$ is a binary mask indicating known and unknown regions in $z$. 

A naive way to use the mask-based guidance in flow matching is to fuse each time-dependent sample $x_t$ before the flow matching update (\cref{eq:fm_sampling}). 
This yields a modified update equation
\begin{equation}\label{eq:cond_gen}
x_{t+\Delta_t} = x_{t}' + \Delta_t v_{\theta,t}(x_{t}'),
\end{equation}
where $x_t'$ is the sample fused according to \cref{eq:fusion}. However, this approach leads to visible misalignments between known and unknown regions in the updated samples $x_{t+\Delta_t}$, particularly at mask boundaries. 
Thus, the distribution of fused samples diverges from the
flow matching model training distribution.
In other words, in this naive strategy, $x_t'$ is unlikely to lie on a trajectory consistent with the learned generative process. 
Consequently, the flow matching model cannot be relied on to map $x'_t$ to a high-quality image.

\paragraph{Trajectory correction}
To address this issue and improve the quality of generated samples, we introduce a trajectory correction mechanism. After the initial mask-guided update (\cref{eq:cond_gen}), we perform a forward extrapolation using the learned velocity field to project the sample toward the endpoint at $t=1$ of the generative path:
\begin{equation}
    \widetilde{x}_{1} = x_{t+\Delta_t} + (1-(t+\Delta_t)) \, v_{\theta, t+\Delta_t}(x_{t+\Delta_t}).
\end{equation}
This extrapolation acts as a learned denoiser, helping to project the sample closer to the data manifold and correct misalignments caused by the mask fusion.

To place the sample at the correct location along the generative trajectory,
we scale the extrapolated sample and reintroduce noise:
\begin{equation}
    x_{t} = t \widetilde{x}_{1} + (1 - t) \eta, \quad \eta \sim \mathcal{N}(0, I).
\end{equation}
This reintroduction of noise adds stochasticity back into the generative process, allowing the flow model to generate diverse and realistic images.

Repeated application of this correction enhances consistency with the known input while allowing the unknown regions to evolve naturally under the flow matching prior. Empirically, we find that even a single correction per ODE step significantly improves the alignment between the restored and observed content (see \cref{fig:qualitative_correction_steps}).

\begin{algorithm}[h]
    \caption{Mask-Guided Restora-Flow Sampling 
    } \label{alg:restora-flow}
    \begin{algorithmic}[1]
        \State \textbf{Input:} learned flow network $v_{\theta}$, degraded observation $z \in \mathbb{R}^d$, number of ODE steps $N$ (with $\Delta_t \gets \frac{1}{N}$), number of corrections $C > 0$, mask $m$
        \State Sample $x \sim \mathcal{N}(0, I)$ 
        \For{$t = 0, \Delta_t, \dots, 1-\Delta_t$}
            \For{$c = 0, \dots, C$}
                \State Sample $\epsilon \sim \mathcal{N}(0, I)$
                \State $z' \gets t z + (1 - t)\epsilon$ \text{ if } $t > 0$, \text{ else } $z' = 0$ 
                \State $x' \gets m \odot z' + (1-m) \odot x$ 
                \State $x \gets x' + \Delta_t \, v_{\theta, t}(x',t)$ 
                 \If{$c > 0$ \textbf{and} $t < 1 - \Delta_t$}
                    \State Sample $\eta \sim \mathcal{N}(0, I)$
                    \State $\widetilde{x}_{1} \gets x + (1-(t+\Delta_t)) \, v_{\theta, t+\Delta_t}(x,t+\Delta_t)$ 
                    \State $x \gets t \widetilde{x}_{1} + (1 - t) \eta$ 
                \Else
                \State $t \gets t + \Delta_t$ 
                \EndIf
            \EndFor
        \EndFor
        \State \Return $x$
    \end{algorithmic}
\end{algorithm}

\begin{algorithm}[h]
    \caption{Restora-Flow Sampling for Denoising} \label{alg:restora-flow_denoise}
    \begin{algorithmic}[1]
        \State \textbf{Input:} degraded observation $z \in \mathbb{R}^d$ with noise level $\sigma$, number of ODE steps $N$ (with $\Delta_t \gets \frac{1}{N}$)
        \State Sample $x_0 \sim \mathcal{N}(0, I)$ 
        \For{$t = 0, \Delta_t, \dots, 1-\Delta_t$}
                \State Sample $\epsilon \sim \mathcal{N}(0, I)$
                \State $z' \gets (1 - \sigma) z$ 
                \State $x_{t+\Delta_t}' \gets x_{t} + \Delta_t \, v_{\theta, t}(x_t)$ 
                \State $x_{t+\Delta_t} \gets \mathbf{1}_{\{t < 1-\sigma\}} z' + \mathbf{1}_{\{t \geq 1-\sigma\}}  x'_{t+\Delta_t}$     
            \EndFor
        \State \Return $x_1$
    \end{algorithmic}
\end{algorithm}

Restora-Flow is summarized in \cref{alg:restora-flow} for mask-based image restoration tasks, such as inpainting and super-resolution. The unconditional sampling process is guided by evolving the sample according to the flow matching prior and applying the mask-guided fusion to incorporate the information of the degraded observation \(z\). This is followed by our proposed correction mechanism for enhanced alignment between generated content and the observation $z$.

\cref{alg:restora-flow_denoise} describes Restora-Flow for image denoising. 
In contrast to \cref{alg:restora-flow}, a time-dependent and global mask \( m(t) \) is used here where we set \( m(t)=1 \) if \( t < 1-\sigma \), else it is 0, as given by the indicator function in \cref{alg:restora-flow_denoise}. 
Thus, the noisy observation \(z\) is used only as an initialization for the sampling process, and for \(t \geq 1-\sigma\) the ODE evolves the solution without further influence from \(z\).

\begin{figure*}[t]
  \centering
    \includegraphics[width=0.99\linewidth]{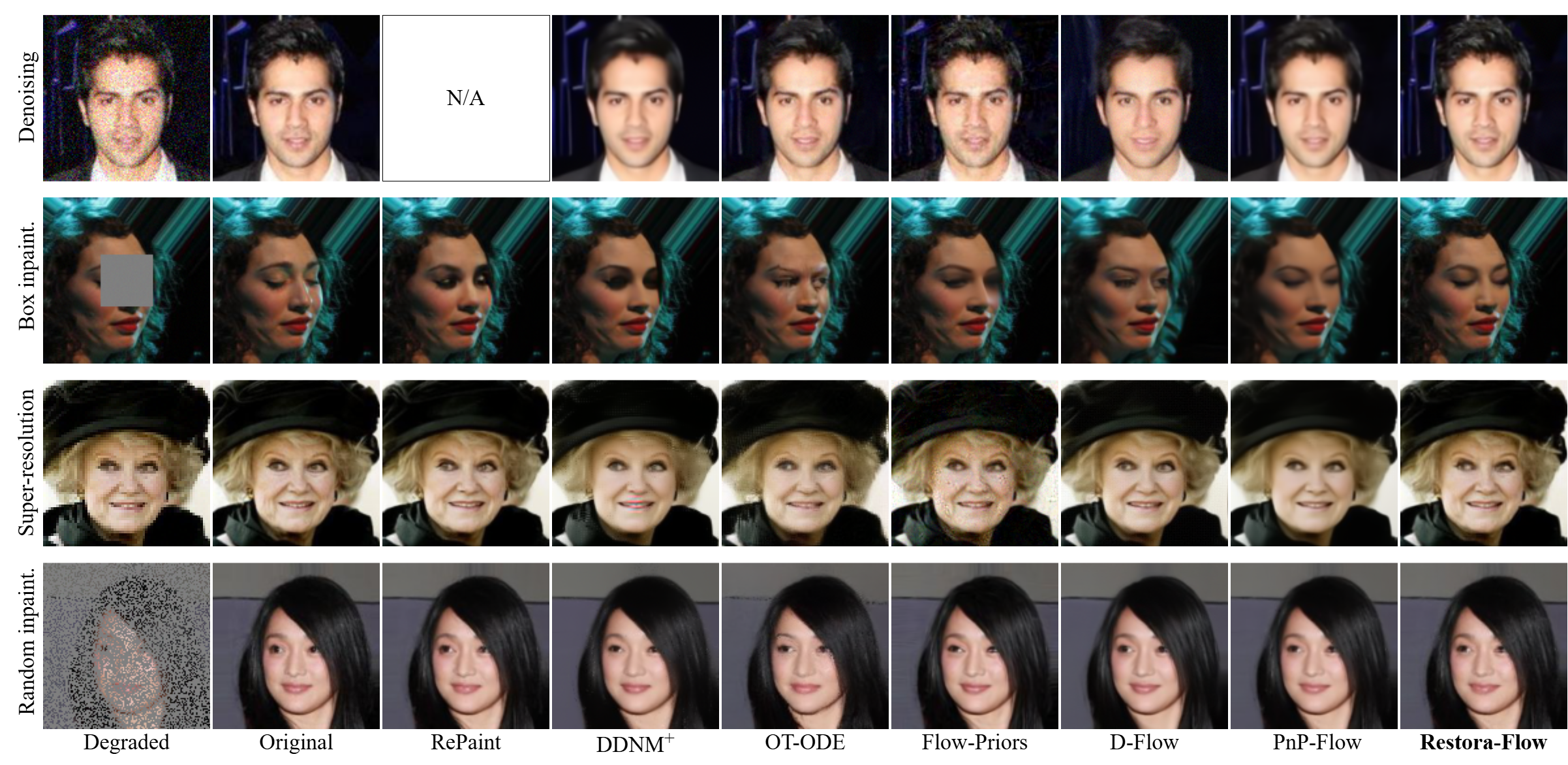}
    \caption{
    Qualitative results on CelebA.
    Shown are the degraded image (col 1), original image (col 2), restored images using related work as indicated (cols 3-8) and restored image of Restora-Flow (col 9).
    Rows refer to denoising (row 1), box inpainting (row 2), super-resolution (row 3) and random inpainting (row 4).
    RePaint is not applicable to denoising (N/A).
    Differences can be best seen in the pdf version.
    }
   \label{fig:celeba_results}
\end{figure*}

\begin{figure*}[t]
    \begin{minipage}{.32\textwidth}
        \centering
        \definecolor{repaint}{HTML}{955196}
\definecolor{otode}{HTML}{87A96B}
\definecolor{flowpriors}{HTML}{CFB53B}
\definecolor{dflow}{HTML}{FF9966}
\definecolor{pnpflow}{HTML}{6699CC}
\definecolor{restoraflow}{HTML}{CC3333}

\definecolor{denoising}{HTML}{9466FF}
\definecolor{boxinpainting}{HTML}{69C7F8}
\definecolor{superresolution}{HTML}{D93F50}
\definecolor{randominpainting}{HTML}{FF9966}

\def\myopacity{1}

\begin{tikzpicture}[scale=0.59]
    \begin{groupplot}[
        group style={
            group name=group,
            group size=2 by 1,
            horizontal sep=10pt,
        },
        ymin=0, ymax=0.15,
        yticklabel style={/pgf/number format/fixed, /pgf/number format/precision=5},
        height=6cm,
        grid=major,
    ]

    \nextgroupplot[xmax=8,
        xtick={0, 2, 4, 6, 8},
        xticklabels={0, 2, 4, 6},
        ytick={0, 0.05, 0.1, 0.15},
        axis y line*=left,
        width=6cm,
        mark options={scale=1.5}
    ]

    \addplot[only marks, mark=+, mark size=1.75, color=denoising, opacity=\myopacity] coordinates {(11.57, 0.076)};
    \addplot[only marks, mark=*, mark size=1.75, color=denoising, opacity=\myopacity] coordinates {(2.95, 0.033)};
    \addplot[only marks, mark=square*, mark size=1.5, color=denoising, opacity=\myopacity] coordinates {(26.22, 0.132)};
    \addplot[only marks, mark=diamond*, color=denoising, opacity=\myopacity] coordinates {(22.73, 0.099)};
    \addplot[only marks, mark=triangle*, color=denoising, opacity=\myopacity] coordinates {(4.60, 0.056)};

    \addplot[only marks, mark=+, mark size=1.75, color=boxinpainting, opacity=\myopacity] coordinates {(11.57, 0.019)};
    \addplot[only marks, mark=asterisk, color=boxinpainting, opacity=\myopacity] coordinates {(32.89, 0.016)};
    \addplot[only marks, mark=*, mark size=1.75, color=boxinpainting] coordinates {(3.68, 0.022)};
    \addplot[only marks, mark=square*, mark size=1.5, color=boxinpainting, opacity=\myopacity] coordinates {(26.22, 0.020)};
    \addplot[only marks, mark=diamond*, color=boxinpainting, opacity=\myopacity] coordinates {(65.81, 0.041)};
    \addplot[only marks, mark=triangle*, color=boxinpainting, opacity=\myopacity] coordinates {(4.60, 0.045)};

    \addplot[only marks, mark=+, mark size=1.75, color=superresolution, opacity=\myopacity] coordinates {(11.57, 0.046)};
    \addplot[only marks, mark=asterisk, color=superresolution, opacity=\myopacity] coordinates {(32.89, 0.014)};
    \addplot[only marks, mark=*, mark size=1.75, color=superresolution, opacity=\myopacity] coordinates {(3.76, 0.055)};
    \addplot[only marks, mark=square*, mark size=1.5, color=superresolution, opacity=\myopacity] coordinates {(26.22, 0.110)};
    \addplot[only marks, mark=diamond*, color=superresolution, opacity=\myopacity] coordinates {(71.43, 0.031)};
    \addplot[only marks, mark=triangle*, color=superresolution, opacity=\myopacity] coordinates {(4.60, 0.058)};
    
    \addplot[only marks, mark=+, mark size=1.75, color=randominpainting, opacity=\myopacity] coordinates {(11.57, 0.031)};
    \addplot[only marks, mark=asterisk, color=randominpainting, opacity=\myopacity] coordinates {(32.89, 0.014)};
    \addplot[only marks, mark=*, mark size=1.75, color=randominpainting, opacity=\myopacity] coordinates {(3.76, 0.051)};
    \addplot[only marks, mark=square*, mark size=1.5, color=randominpainting, opacity=\myopacity] coordinates {(26.22, 0.019)};
    \addplot[only marks, mark=diamond*, color=randominpainting, opacity=\myopacity] coordinates {(131.78, 0.021)}; 
    \addplot[only marks, mark=triangle*, color=randominpainting, opacity=\myopacity] coordinates {(4.60, 0.022)};
    
    \addplot[only marks, mark=pentagon*, mark options={draw=black, line width=0.05mm, fill=denoising, opacity=\myopacity}, mark size=4] coordinates {(0.58, 0.019)};
    \addplot[only marks, mark=pentagon*, mark options={draw=black, line width=0.05mm, fill=boxinpainting, opacity=\myopacity}, mark size=4] coordinates {(2.06, 0.018)};
    \addplot[only marks, mark=pentagon*, mark options={draw=black, line width=0.05mm, fill=superresolution, opacity=\myopacity}, mark size=4] coordinates {(3.63, 0.014)};
    \addplot[only marks, mark=pentagon*, mark options={draw=black, line width=0.05mm, fill=randominpainting, opacity=\myopacity}, mark size=4] coordinates {(3.63, 0.015)};

    \nextgroupplot[xmin=8,xmax=150,
        xtick={8, 50, 100, 150},
        ylabel={},
        ytick={},
        yticklabels={},
        axis y line*=right,
        width=4cm,
        mark options={scale=1.5}
    ]

    \addplot[only marks, mark=+, mark size=1.75, color=denoising, opacity=\myopacity] coordinates {(11.57, 0.076)};
    \addplot[only marks, mark=*, mark size=1.75, color=denoising, opacity=\myopacity] coordinates {(2.95, 0.033)};
    \addplot[only marks, mark=square*, mark size=1.5, color=denoising, opacity=\myopacity] coordinates {(26.22, 0.132)};
    \addplot[only marks, mark=diamond*, color=denoising, opacity=\myopacity] coordinates {(22.73, 0.099)};
    \addplot[only marks, mark=triangle*, color=denoising, opacity=\myopacity] coordinates {(4.60, 0.056)};

    \addplot[only marks, mark=+, mark size=1.75, color=boxinpainting, opacity=\myopacity] coordinates {(11.57, 0.019)};
    \addplot[only marks, mark=asterisk, color=boxinpainting, opacity=\myopacity] coordinates {(32.89, 0.016)};
    \addplot[only marks, mark=*, mark size=1.75, color=boxinpainting] coordinates {(3.68, 0.022)};
    \addplot[only marks, mark=square*, mark size=1.5, color=boxinpainting, opacity=\myopacity] coordinates {(26.22, 0.020)};
    \addplot[only marks, mark=diamond*, color=boxinpainting, opacity=\myopacity] coordinates {(65.81, 0.041)};
    \addplot[only marks, mark=triangle*, color=boxinpainting, opacity=\myopacity] coordinates {(4.60, 0.045)};

    \addplot[only marks, mark=+, mark size=1.75, color=superresolution, opacity=\myopacity] coordinates {(11.57, 0.046)};
    \addplot[only marks, mark=asterisk, color=superresolution, opacity=\myopacity] coordinates {(32.89, 0.014)};
    \addplot[only marks, mark=*, mark size=1.75, color=superresolution, opacity=\myopacity] coordinates {(3.76, 0.055)};
    \addplot[only marks, mark=square*, mark size=1.5, color=superresolution, opacity=\myopacity] coordinates {(26.22, 0.110)};
    \addplot[only marks, mark=diamond*, color=superresolution, opacity=\myopacity] coordinates {(71.43, 0.031)};
    \addplot[only marks, mark=triangle*, color=superresolution, opacity=\myopacity] coordinates {(4.60, 0.058)};
    
    \addplot[only marks, mark=+, mark size=1.75, color=randominpainting, opacity=\myopacity] coordinates {(11.57, 0.031)};
    \addplot[only marks, mark=asterisk, color=randominpainting, opacity=\myopacity] coordinates {(32.89, 0.014)};
    \addplot[only marks, mark=*, mark size=1.75, color=randominpainting, opacity=\myopacity] coordinates {(3.76, 0.051)};
    \addplot[only marks, mark=square*, mark size=1.5, color=randominpainting, opacity=\myopacity] coordinates {(26.22, 0.019)};
    \addplot[only marks, mark=diamond*, color=randominpainting, opacity=\myopacity] coordinates {(131.78, 0.021)}; 
    \addplot[only marks, mark=triangle*, color=randominpainting, opacity=\myopacity] coordinates {(4.60, 0.022)};
    
    \addplot[only marks, mark=pentagon*, mark options={draw=black, line width=0.05mm, fill=denoising, opacity=\myopacity}, mark size=4] coordinates {(0.58, 0.019)};
    \addplot[only marks, mark=pentagon*, mark options={draw=black, line width=0.05mm, fill=boxinpainting, opacity=\myopacity}, mark size=4] coordinates {(2.06, 0.018)};
    \addplot[only marks, mark=pentagon*, mark options={draw=black, line width=0.05mm, fill=superresolution, opacity=\myopacity}, mark size=4] coordinates {(3.63, 0.014)};
    \addplot[only marks, mark=pentagon*, mark options={draw=black, line width=0.05mm, fill=randominpainting, opacity=\myopacity}, mark size=4] coordinates {(3.63, 0.015)};

    \end{groupplot}

    \node (title) at ($(group c1r1.north)!0.5!(group c2r1.north)+(-0.8,0.5cm)$) {\scriptsize CelebA: LPIPS $\downarrow$};
    \node (xlabel) at ($(group c1r1.south)!0.5!(group c2r1.south)+(-0.8,-0.8cm)$) {\scriptsize Time (s)};
    \node[rotate=90] (ylabel) at ($(group c1r1.west)+(-1.2cm,0)$) {\scriptsize Score};

    \node[inner sep=0, anchor=north west] (tilde) at ($(group c1r1.east)+(-0.01cm,0)$) {\tiny $\sim$};
    
\end{tikzpicture}
    \end{minipage}
    \hspace{2pt}
    \begin{minipage}{.32\textwidth}
        \centering
        \definecolor{repaint}{HTML}{955196}
\definecolor{otode}{HTML}{a4d4ff}
\definecolor{flowpriors}{HTML}{CFB53B}
\definecolor{dflow}{HTML}{FF9966}
\definecolor{pnpflow}{HTML}{6699CC}
\definecolor{restoraflow}{HTML}{CC3333}

\definecolor{denoising}{HTML}{9466FF}
\definecolor{boxinpainting}{HTML}{69C7F8}
\definecolor{superresolution}{HTML}{D93F50}
\definecolor{randominpainting}{HTML}{FF9966}

\def\myopacity{1}

\begin{tikzpicture}[scale=0.59]
    \begin{groupplot}[
        group style={
            group name=group,
            group size=2 by 1,
            horizontal sep=10pt
        },
        ymin=0.6, ymax=1,
        yticklabel style={/pgf/number format/fixed, /pgf/number format/precision=5},
        height=6cm,
        grid=major,
    ]

    \nextgroupplot[xmax=8,
        xtick={0, 2, 4, 6, 8},
        xticklabels={0, 2, 4, 6},
        ytick={0.6, 0.7, 0.8, 0.9, 1},
        axis y line*=left,
        width=6cm,
        mark options={scale=1.5}
    ]
       
    \addplot[only marks, mark=+, mark size=1.75, color=denoising, opacity=\myopacity] coordinates {(11.57, 0.885)};
    \addplot[only marks, mark=*, mark size=1.75, color=denoising, opacity=\myopacity] coordinates {(2.95, 0.858)};
    \addplot[only marks, mark=square*, mark size=1.5, color=denoising, opacity=\myopacity] coordinates {(26.22, 0.767)};
    \addplot[only marks, mark=diamond*, color=denoising, opacity=\myopacity] coordinates {(22.73, 0.695)};
    \addplot[only marks, mark=triangle*, color=denoising, opacity=\myopacity] coordinates {(4.60, 0.910)};
    
    \addplot[only marks, mark=+, mark size=1.75, color=boxinpainting, opacity=\myopacity] coordinates {(11.57, 0.969)};
    \addplot[only marks, mark=asterisk, color=boxinpainting, opacity=\myopacity] coordinates {(32.89, 0.967)};
    \addplot[only marks, mark=*, mark size=1.75, color=boxinpainting, opacity=\myopacity] coordinates {(3.68, 0.954)};
    \addplot[only marks, mark=square*, mark size=1.5, color=boxinpainting, opacity=\myopacity] coordinates {(26.22, 0.969)};
    \addplot[only marks, mark=diamond*, color=boxinpainting, opacity=\myopacity] coordinates {(65.81, 0.907)};
    \addplot[only marks, mark=triangle*, color=boxinpainting, opacity=\myopacity] coordinates {(4.60, 0.941)};
    
    \addplot[only marks, mark=+, mark size=1.75, color=superresolution, opacity=\myopacity] coordinates {(11.57, 0.905)};
    \addplot[only marks, mark=asterisk, color=superresolution, opacity=\myopacity] coordinates {(32.89, 0.946)};
    \addplot[only marks, mark=*, mark size=1.75, color=superresolution, opacity=\myopacity] coordinates {(3.76, 0.870)};
    \addplot[only marks, mark=square*, mark size=1.5, color=superresolution, opacity=\myopacity] coordinates {(26.22, 0.722)};
    \addplot[only marks, mark=diamond*, color=superresolution, opacity=\myopacity] coordinates {(71.43, 0.894)};
    \addplot[only marks, mark=triangle*, color=superresolution, opacity=\myopacity] coordinates {(4.60, 0.908)};
    
    \addplot[only marks, mark=+, mark size=1.75, color=randominpainting, opacity=\myopacity] coordinates {(11.57, 0.920)};
    \addplot[only marks, mark=asterisk, color=randominpainting, opacity=\myopacity] coordinates {(32.89, 0.945)};
    \addplot[only marks, mark=*, mark size=1.75, color=randominpainting, opacity=\myopacity] coordinates {(3.76, 0.871)};
    \addplot[only marks, mark=square*, mark size=1.5, color=randominpainting, opacity=\myopacity] coordinates {(26.22, 0.944)};
    \addplot[only marks, mark=diamond*, color=randominpainting, opacity=\myopacity] coordinates {(131.78, 0.931)}; 
    \addplot[only marks, mark=triangle*, color=randominpainting, opacity=\myopacity] coordinates {(4.60, 0.954)};
    
    \addplot[only marks, mark=pentagon*, mark options={draw=black, line width=0.05mm, fill=denoising, opacity=\myopacity}, mark size=4] coordinates {(0.58, 0.922)};
    \addplot[only marks, mark=pentagon*, mark options={draw=black, line width=0.05mm, fill=boxinpainting, opacity=\myopacity}, mark size=4] coordinates {(2.06, 0.964)};
    \addplot[only marks, mark=pentagon*, mark options={draw=black, line width=0.05mm, fill=superresolution, opacity=\myopacity}, mark size=4] coordinates {(3.63, 0.952)};
    \addplot[only marks, mark=pentagon*, mark options={draw=black, line width=0.05mm, fill=randominpainting, opacity=\myopacity}, mark size=4] coordinates {(3.63, 0.947)};

    \nextgroupplot[xmin=8,xmax=150,
        xtick={8, 50, 100, 150},
        ylabel={},
        ytick={},
        yticklabels={},
        axis y line*=right,
        width=4cm,
        mark options={scale=1.5}
    ]
    \addplot[only marks, mark=diamond*, color=randominpainting] coordinates {(131.78, 0.931)};

    \addplot[only marks, mark=+, mark size=1.75, color=denoising, opacity=\myopacity] coordinates {(11.57, 0.885)};
    \addplot[only marks, mark=*, mark size=1.75, color=denoising, opacity=\myopacity] coordinates {(2.95, 0.858)};
    \addplot[only marks, mark=square*, mark size=1.5, color=denoising, opacity=\myopacity] coordinates {(26.22, 0.767)};
    \addplot[only marks, mark=diamond*, color=denoising, opacity=\myopacity] coordinates {(22.73, 0.695)};
    \addplot[only marks, mark=triangle*, color=denoising, opacity=\myopacity] coordinates {(4.60, 0.910)};
    
    \addplot[only marks, mark=+, mark size=1.75, color=boxinpainting, opacity=\myopacity] coordinates {(11.57, 0.969)};
    \addplot[only marks, mark=asterisk, color=boxinpainting, opacity=\myopacity] coordinates {(32.89, 0.967)};
    \addplot[only marks, mark=*, mark size=1.75, color=boxinpainting, opacity=\myopacity] coordinates {(3.68, 0.954)};
    \addplot[only marks, mark=square*, mark size=1.5, color=boxinpainting, opacity=\myopacity] coordinates {(26.22, 0.969)};
    \addplot[only marks, mark=diamond*, color=boxinpainting, opacity=\myopacity] coordinates {(65.81, 0.907)};
    \addplot[only marks, mark=triangle*, color=boxinpainting, opacity=\myopacity] coordinates {(4.60, 0.941)};
    
    \addplot[only marks, mark=+, mark size=1.75, color=superresolution, opacity=\myopacity] coordinates {(11.57, 0.905)};
    \addplot[only marks, mark=asterisk, color=superresolution, opacity=\myopacity] coordinates {(32.89, 0.946)};
    \addplot[only marks, mark=*, mark size=1.75, color=superresolution, opacity=\myopacity] coordinates {(3.76, 0.870)};
    \addplot[only marks, mark=square*, mark size=1.5, color=superresolution, opacity=\myopacity] coordinates {(26.22, 0.722)};
    \addplot[only marks, mark=diamond*, color=superresolution, opacity=\myopacity] coordinates {(71.43, 0.894)};
    \addplot[only marks, mark=triangle*, color=superresolution, opacity=\myopacity] coordinates {(4.60, 0.908)};
    
    \addplot[only marks, mark=+, mark size=1.75, color=randominpainting, opacity=\myopacity] coordinates {(11.57, 0.920)};
    \addplot[only marks, mark=asterisk, color=randominpainting, opacity=\myopacity] coordinates {(32.89, 0.945)};
    \addplot[only marks, mark=*, mark size=1.75, color=randominpainting, opacity=\myopacity] coordinates {(3.76, 0.871)};
    \addplot[only marks, mark=square*, mark size=1.5, color=randominpainting, opacity=\myopacity] coordinates {(26.22, 0.944)};
    \addplot[only marks, mark=diamond*, color=randominpainting, opacity=\myopacity] coordinates {(131.78, 0.931)}; 
    \addplot[only marks, mark=triangle*, color=randominpainting, opacity=\myopacity] coordinates {(4.60, 0.954)};
    
    \addplot[only marks, mark=pentagon*, mark options={draw=black, line width=0.05mm, fill=denoising, opacity=\myopacity}, mark size=4] coordinates {(0.58, 0.922)};
    \addplot[only marks, mark=pentagon*, mark options={draw=black, line width=0.05mm, fill=boxinpainting, opacity=\myopacity}, mark size=4] coordinates {(2.06, 0.964)};
    \addplot[only marks, mark=pentagon*, mark options={draw=black, line width=0.05mm, fill=superresolution, opacity=\myopacity}, mark size=4] coordinates {(3.63, 0.952)};
    \addplot[only marks, mark=pentagon*, mark options={draw=black, line width=0.05mm, fill=randominpainting, opacity=\myopacity}, mark size=4] coordinates {(3.63, 0.947)};
    
    \end{groupplot}

    \node (title) at ($(group c1r1.north)!0.5!(group c2r1.north)+(-0.8,0.5cm)$) {\scriptsize CelebA: SSIM $\uparrow$};
    \node (title) at ($(group c1r1.south)!0.5!(group c2r1.south)+(-0.8,-0.8cm)$) {\scriptsize Time (s)};
    \node[rotate=90] (ylabel) at ($(group c1r1.west)+(-1.2cm,0)$) {\scriptsize Score};

    \node[inner sep=0, anchor=north west] (tilde) at ($(group c1r1.east)+(-0.01cm,0)$) {\tiny $\sim$};

\end{tikzpicture}
    \end{minipage}
    \hspace{2pt}
    \begin{minipage}{.32\textwidth}
        \centering
        \input{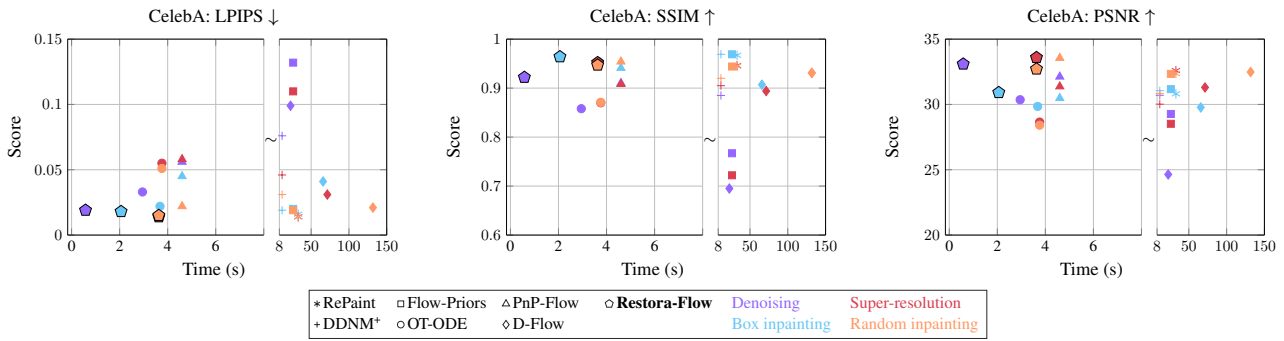}
    \end{minipage}

        \centering
        \definecolor{repaint}{HTML}{955196}
\definecolor{otode}{HTML}{87A96B}
\definecolor{flowpriors}{HTML}{CFB53B}
\definecolor{dflow}{HTML}{FF9966}
\definecolor{pnpflow}{HTML}{6699CC}
\definecolor{restoraflow}{HTML}{CC3333}

\definecolor{denoising}{HTML}{9466FF}
\definecolor{boxinpainting}{HTML}{69C7F8}
\definecolor{superresolution}{HTML}{D93F50}
\definecolor{randominpainting}{HTML}{FF9966}

    



\begin{tikzpicture}[scale=0.63]
    \begin{axis}[%
    legend columns=6, 
    hide axis,
    xmin=10,
    xmax=100,
    ymin=0,
    ymax=0.4,
    height=2cm,
    width=10cm,
    legend style={draw=white!15!black, legend cell align=left}
    ]
    

    \addlegendimage{mark=asterisk,only marks,mark size=2}
    \addlegendentry{RePaint \phantom{x}}
    \addlegendimage{mark=square,only marks,mark size=1.75}
    \addlegendentry{Flow-Priors \phantom{x}}
    \addlegendimage{mark=triangle,only marks,mark size=2.5}
    \addlegendentry{PnP-Flow \phantom{x}}
    \addlegendimage{mark=pentagon,only marks,mark size=2.5}
    \addlegendentry{\textbf{Restora-Flow} \phantom{x}}
    \addlegendimage{empty legend}
    \addlegendentry[color=denoising]{Denoising \phantom{x}}
    \addlegendimage{empty legend}
    \addlegendentry[color=superresolution]{Super-resolution \phantom{x}}
    \addlegendimage{mark=+,only marks,mark size=1.75}
    \addlegendentry{DDNM\textsuperscript{+} \phantom{x}}
    \addlegendimage{mark=o,only marks,mark size=2}
    \addlegendentry{OT-ODE \phantom{x}}
    \addlegendimage{mark=diamond,only marks,mark size=2.5}
    \addlegendentry{D-Flow \phantom{x}}
    \addlegendimage{empty legend}
    \addlegendentry[color=boxinpainting]{\phantom{x}}
    \addlegendimage{empty legend}
    \addlegendentry[color=boxinpainting]{Box inpainting \phantom{x}}
    \addlegendimage{empty legend}
    \addlegendentry[color=randominpainting]{Random inpainting}
    
    \end{axis}
\end{tikzpicture}  
    \caption{
    Visual representation of quantitative results on CelebA.
    Restora-Flow (\pentagon) is compared to related work methods (other shapes) on four different tasks (colors).
    The plots show LPIPS~$\downarrow$ (left), SSIM~$\uparrow$ (center) and PSNR~$\uparrow$ (right) on the y-axis, and processing time~$\downarrow$ (all plots) on the x-axis.
    For better visualization and comparison, each plot is separated into two parts with different scales in the x-axis.
    }
    \label{fig:performance_vs_time_celeba}
\end{figure*}

\section{Experimental Setup}
\label{sec:experimental_setup}
\paragraph{Datasets.} 
To assess the performance of Restora-Flow, we utilize four datasets: (1)~CelebA~\cite{liu2015faceattributes}, featuring 162k training images of celebrity faces resized to $128\times128$ pixels; (2)~AFHQ-Cat~\cite{choi2020stargan}, which includes 5k training images of cat faces resized to $256\times256$ pixels; (3)~COCO~\cite{lin2014microsoft}, containing 118k training images of various object types resized to $128\times128$ pixels; and (4)~X-ray Hand~\cite{gertych2007bone,zhang2009racial}, comprising 895 hand radiographs resized to $256\times256$ pixels. In line with recent work \cite{martin2024pnp}, we report results on the same 100 test images from the CelebA dataset and on the same 100 test images from the AFHQ-Cat dataset. Moreover, we report results on 100 validation images from the COCO dataset and 298 test images from the medical X-ray Hand dataset.

\paragraph{Tasks.} We report results for 
denoising, box inpainting, super-resolution and random inpainting. Additionally, we include results for a clinically motivated occlusion removal task on the X-ray Hand dataset.
All experiments, unless otherwise noted, include a Gaussian measurement noise level of \(\sigma = 0.01\).
For the CelebA and COCO datasets, we employ denoising with a noise level of \(\sigma = 0.2\), box inpainting with a \(40\times40\) centered mask, \(2\times\) super-resolution, and random inpainting with a mask covering \(70\%\) of the pixels. 
For the AFHQ-Cat dataset, we perform denoising with a noise level of \(\sigma = 0.2\), box inpainting using an \(80\times80\) centered mask, $4\times$ super-resolution, and random inpainting with a mask covering \(70\%\) of the pixels. 
For the X-ray Hand dataset, we employ denoising with \(\sigma = 0.08\), box inpainting with a \(128\times128\) centered mask, and $2\times$ super-resolution. For occlusion removal, we synthetically add occlusions to images as described in~\cite{hadzic2025flow}.

\paragraph{Baselines and Metrics.}
We compare Restora-Flow with four flow-based baselines: OT-ODE~\cite{pokle2024training}, Flow-Priors~\cite{zhang2024flow}, D-Flow~\cite{benhamu2024dflow} and PnP-Flow~\cite{martin2024pnp} (see \cref{sec:related_work}).
In our experiments on CelebA and AFHQ-Cat, we adopt the hyperparameter values defined in~\cite{martin2024pnp}, where a grid search was performed to identify optimal settings for each baseline on the respective datasets. 
Specifically, OT-ODE, Flow-Priors and PnP-Flow each require tuning two hyperparameters, while Flow-Priors requires tuning three hyperparameters, with different values chosen based on the task and dataset. For the COCO dataset, we use the same hyperparameters as for CelebA. For the X-ray Hand dataset, we adopt the same hyperparameters as those used for AFHQ-Cat.

Furthermore, we compare Restora-Flow with diffusion-based methods, \ie, RePaint~\cite{lugmayr2022repaint} and DDNM\textsuperscript{+}~\cite{wang2023zeroshot}. 
For RePaint, we use their proposed hyperparameter values~\cite{lugmayr2022repaint}: a jump length of 10 and 10 resampling steps. For DDNM\textsuperscript{+}, we achieved optimal performance using the time-travel trick \cite{lugmayr2022repaint,wang2023zeroshot} parameters $s=1$ and $l=5$.

Following prior work~\cite{pokle2024training,zhang2024flow,benhamu2024dflow,martin2024pnp}, 
we utilize pixel-based Structural Similarity Index (SSIM) and Peak Signal-to-Noise Ratio (PSNR) as distortion metrics. Given the inherent perception-distortion trade-off~\cite{blau2018perception}, we also include Learned Perceptual Image Patch Similarity (LPIPS)~\cite{zhang2018unreasonable} as perceptual metric. Furthermore, we report processing time in seconds to restore a single image.

\begin{table*}[t]
    \caption{
    Quantitative results on CelebA, AFHQ-Cat, COCO and X-ray Hand (from top to bottom).
    Shown are LPIPS~$\downarrow$, SSIM~$\uparrow$, PSNR~$\uparrow$ and processing time~$\downarrow$ for denoising, box inpainting, super-resolution and random inpainting.
    For X-ray Hand, random inpainting is replaced by occlusion removal.
    Restora-Flow is compared to several state-of-the-art methods.
    Best and second best scores are highlighted using bold or underline, respectively.
    RePaint is not applicable to denoising (N/A).
    }
    \centering   
    \label{tab:combined_results}
    \resizebox{\textwidth}{!}{%
    \begin{tabular}{l|cccr|cccr|cccr|cccr}
    \multicolumn{16}{c}{\textbf{CelebA}} \\
    \hline   
        \multirow{2}{*}{Model} & \multicolumn{3}{c}{Denoising $\sigma=0.2$} & Time & \multicolumn{3}{c}{Box inpainting $40\times40$} & Time & \multicolumn{3}{c}{Super-resolution \(2\times\)} & Time & \multicolumn{3}{c}{Random inpainting $70\%$} & Time \\ 
        & LPIPS & SSIM & PSNR & in s & LPIPS & SSIM & PSNR & in s & LPIPS & SSIM & PSNR & in s & LPIPS & SSIM & PSNR & in s\\
        \hline
        
        RePaint~\cite{lugmayr2022repaint} & N/A & N/A & N/A & N/A & \textbf{0.016} & \underline{0.967} & 30.81 & 32.89 & \textbf{0.014} & \underline{0.946} & \underline{32.59} & 32.89 & \textbf{0.014} & 0.945 & 32.37 & 32.89 \\  
        
        DDNM\textsuperscript{+}~\cite{wang2023zeroshot} & 0.076 & 0.885 & 30.70 & 11.57 & 0.019 & \textbf{0.969} & \underline{31.05} & 11.57 & 0.046 & 0.905 & 30.02 & 11.57 & 0.031 & 0.920 & 30.83 & 11.57 \\  
        
        OT-ODE~\cite{pokle2024training} & \underline{0.033} & 0.858 & 30.36 & \underline{2.95} & 0.022 & 0.954 & 29.85 & \underline{3.68} & 0.055 & 0.870 & 28.65 & \underline{3.76} & 0.051 & 0.871 & 28.41 & \underline{3.76} \\   
        
        Flow-Priors~\cite{zhang2024flow} & 0.132 & 0.767 & 29.27 & 26.22 & 0.020 & \textbf{0.969}  & \textbf{31.17} & 26.22 & 0.110 & 0.722 & 28.52 & 26.22 & 0.019 & 0.944 & 32.34 & 26.22 \\  
        
        D-Flow~\cite{benhamu2024dflow} & 0.099 & 0.695 & 24.64 & 22.73 & 0.041 & 0.907 & 29.77 & 65.81 & \underline{0.031} & 0.894 & 31.30 & 71.43 & 0.021 & 0.931 & 32.48 & 131.78 \\    
        
        PnP-Flow~\cite{martin2024pnp} & 0.056 & \underline{0.910} & \underline{32.12} & 4.60 & 0.045 & 0.941 & 30.48 & 4.60 & 0.058 & 0.908 & 31.37 & 4.60 & 0.022 & \textbf{0.954} & \textbf{33.55} & 4.60 \\
        \hline
        
        Restora-Flow & \textbf{0.019} & \textbf{0.922} & \textbf{33.09} & \textbf{0.58} & \underline{0.018} & 0.964 & 30.91 & \textbf{2.06} &  \textbf{0.014} & \textbf{0.952} & \textbf{33.59} & \textbf{3.63} & \underline{0.015} & \underline{0.947} & \underline{32.71} & \textbf{3.63} \\   
    \hline   
    \multicolumn{16}{c}{} \\
    \multicolumn{16}{c}{\textbf{AFHQ-Cat}} \\
    \hline   
        \multirow{2}{*}{Model} & \multicolumn{3}{c}{Denoising $\sigma=0.2$} & Time & \multicolumn{3}{c}{Box inpainting $80\times80$} & Time & \multicolumn{3}{c}{Super-resolution \(4\times\)} & Time & \multicolumn{3}{c}{Random inpainting $70\%$} & Time \\ 
        & LPIPS & SSIM & PSNR & in s & LPIPS & SSIM & PSNR & in s & LPIPS & SSIM & PSNR & in s & LPIPS & SSIM & PSNR & in s\\
        \hline
        RePaint~\cite{lugmayr2022repaint} & N/A & N/A & N/A & N/A & \textbf{0.043} & \underline{0.939} & \textbf{26.26} & 86.23 & \underline{0.139} & 0.701 & 24.71 & 86.23 & \textbf{0.034} & 0.897 & 30.93 & 86.23  \\ 

        DDNM\textsuperscript{+}~\cite{wang2023zeroshot}  & 0.170 & 0.818 & 29.06 & 13.74 & 0.048 & \textbf{0.942} & 25.16 & 13.74 &0.462 & 0.534 & 19.69 & \underline{13.74} & 0.065 & 0.876 & 30.12 & 13.74  \\ 

        OT-ODE~\cite{pokle2024training} & \underline{0.078} & 0.814 & 29.73 & \underline{5.54} & 0.048 & 0.924 & 24.36 & \underline{6.94} & 0.285 & 0.565 & 21.85 & \textbf{7.28} & 0.094 & 0.839 & 28.87 & \textbf{7.28} \\   
        
        Flow-Priors~\cite{zhang2024flow} & 0.153 & 0.771 & 29.43 & 67.10 & 0.054 & \textbf{0.942} & 26.04 & 67.05 & 0.271 & 0.565 & 23.50 & 67.30 & 0.046 & 0.909 & 31.82 & 67.69 \\   
        
        D-Flow~\cite{benhamu2024dflow} & 0.184 & 0.648 & 24.98 & 44.45 & 0.112 & 0.839 & 26.17 & 126.09 & \textbf{0.123} &  0.707 & 25.34 & 261.84 & 0.056 & 0.878 & 30.97 & 266.18 \\  
        
        PnP-Flow~\cite{martin2024pnp} & 0.165 & \underline{0.864} & \underline{31.10} & 9.86 & 0.124 & 0.904 & \underline{26.18} & 9.86 & 0.180 & \textbf{0.790} & \textbf{26.95} & 46.26 & \underline{0.042} & \textbf{0.930} & \textbf{33.07} & 19.15 \\ 
        \hline
        
        Restora-Flow & \textbf{0.051} & \textbf{0.899} & \textbf{32.35} & \textbf{0.72} & \underline{0.047} & \underline{0.939} & 25.96 & \textbf{3.96} & 0.158 & \underline{0.761} & \underline{26.33} & 14.48 &  \textbf{0.034} & \underline{0.914} & \underline{31.99} & \underline{7.48} \\   
    \hline   
    \multicolumn{16}{c}{} \\
    \multicolumn{16}{c}{\textbf{COCO}} \\
    \hline   
        \multirow{2}{*}{Model} & \multicolumn{3}{c}{Denoising $\sigma=0.2$} & Time & \multicolumn{3}{c}{Box inpainting $40\times40$} & Time & \multicolumn{3}{c}{Super-resolution \(2\times\)} & Time & \multicolumn{3}{c}{Random inpainting $70\%$} & Time \\ 
        & LPIPS & SSIM & PSNR & in s & LPIPS & SSIM & PSNR & in s & LPIPS & SSIM & PSNR & in s & LPIPS & SSIM & PSNR & in s\\
        \hline

        RePaint~\cite{lugmayr2022repaint} & N/A & N/A & N/A & N/A & 0.093 & 0.922 & 21.20 & 32.89 & \underline{0.046} & \underline{0.856} & 25.84 & 32.89 & \textbf{0.038} & 0.876 & 26.82 & 32.89 \\ 
        
        DDNM\textsuperscript{+}~\cite{wang2023zeroshot} & 0.162 & 0.805 & 27.04 & 11.57 & 0.112 & 0.925 & 21.71 & 11.57 & 0.257  & 0.682 & 19.05 & 11.57 & 0.069 & 0.845 & 25.80 & 11.57\\  
        
        OT-ODE~\cite{pokle2024training} & \underline{0.066} & 0.810 & 27.52 & \underline{2.95} & \textbf{0.073} & 0.914 & 23.40 & \underline{3.68} & 0.146 & 0.745 & 23.83 & \underline{3.76} & 0.130 & 0.763 & 23.98 & \underline{3.76}  \\   
        
        Flow-Priors~\cite{zhang2024flow} & 0.116 & 0.751 & 27.08 & 26.22 & \underline{0.084} & \underline{0.927} & 23.58 & 26.22 & 0.112 & 0.698 & 24.93 & 26.22 & 0.055 & 0.855 & 25.97 & 26.22 \\ 
        
        D-Flow~\cite{benhamu2024dflow} & 0.252 & 0.552 & 21.19 & 22.73 & 0.115 & 0.825 & 23.46 & 65.81 & 0.083 & 0.778 & 24.80 & 71.43 & 0.053 & 0.840 & 26.29 & 131.78 \\   
        
        PnP-Flow~\cite{martin2024pnp} & 0.128 & \underline{0.855} & \underline{28.97} & 4.60 & 0.121 & 0.892 & \underline{24.56} & 4.60 & 0.118 & 0.827 & \underline{26.73} & 4.60 & 0.053 & \textbf{0.896} & \textbf{28.13} & 4.60  \\   
        
        \hline
        Restora-Flow & \textbf{0.026} & \textbf{0.905} & \textbf{30.57} & \textbf{0.58} & \underline{0.084} & \textbf{0.929} & \textbf{24.80} & \textbf{2.06} & \textbf{0.044} & \textbf{0.877} & \textbf{27.44} & \textbf{3.63} & \underline{0.040} & \underline{0.881} & \underline{27.37} & \textbf{3.63}  \\ 
    \hline   
    \multicolumn{16}{c}{} \\
    \multicolumn{16}{c}{\textbf{X-ray Hand}} \\
    \hline   
        \multirow{2}{*}{Model} & \multicolumn{3}{c}{Denoising $\sigma=0.08$} & Time & \multicolumn{3}{c}{Box inpainting $128\times128$} & Time & \multicolumn{3}{c}{Super-resolution \(2\times\)} & Time & \multicolumn{3}{c}{Occlusion removal} & Time \\ 
        & LPIPS & SSIM & PSNR & in s & LPIPS & SSIM & PSNR & in s & LPIPS & SSIM & PSNR & in s & LPIPS & SSIM & PSNR & in s\\
        \hline

        RePaint~\cite{lugmayr2022repaint} & N/A & N/A & N/A & N/A & 0.046 & 0.821 & 23.90 & 17.02 & 0.074 & 0.767 & 20.04 & 17.02 & 0.032 & 0.898 & \underline{29.66} & 17.02  \\   
        
        DDNM\textsuperscript{+}~\cite{wang2023zeroshot} & 0.057 & 0.819 & 23.78 & 13.35 & 0.059 & 0.801 & 22.76 & 13.35 & 0.143 & 0.635 & 14.10 & 13.35 & 0.047 & 0.884 & 26.57 & 13.35  \\  
        
        OT-ODE~\cite{pokle2024training} & \underline{0.026} & 0.853 & 27.83 & \underline{8.73} & \underline{0.038} & 0.801 & 23.58 & \underline{11.17} & 0.076 & 0.684 & 22.01 & \underline{11.17} & 0.029 & 0.845 & 26.55 & \underline{11.17}  \\

        Flow-Priors~\cite{zhang2024flow} & 0.033 & \underline{0.885} & \underline{28.58} & 68.54 & \textbf{0.035} & \textbf{0.882} & \textbf{25.74} & 68.59 & 0.162 & 0.460 & 20.38 & 68.62 & \underline{0.023} & \underline{0.933} & 27.07 & 68.59  \\  

        D-Flow~\cite{benhamu2024dflow} & 0.077 & 0.630 & 24.09 & 101.66 & 0.145 & 0.588 & 13.61 & 285.55 & 0.127 & 0.639 & 15.26 & 361.22 & 0.110 & 0.587 & 22.23 & 361.22  \\  

        PnP-Flow~\cite{martin2024pnp} & 0.052 & 0.843 & 25.17 & 20.48 & 0.054 & 0.822 & 23.67 & 20.35 &  \textbf{0.029} & \textbf{0.884} & \textbf{25.88} & 102.29 & 0.045 & 0.889 & 26.83 & 20.35  \\  
        
        \hline
        Restora-Flow & \textbf{0.021} & \textbf{0.912} & \textbf{31.34} & \textbf{0.50} & \textbf{0.035} & \underline{0.846} & \underline{24.67} & \textbf{4.03} & \underline{0.037} & \underline{0.857} & \underline{24.66} & \textbf{7.95} & \textbf{0.017} & \textbf{0.935} & \textbf{33.51} & \textbf{4.03} \\ 
    \hline   
    \end{tabular}  
    }
\end{table*}

\begin{figure}[t]
    \input{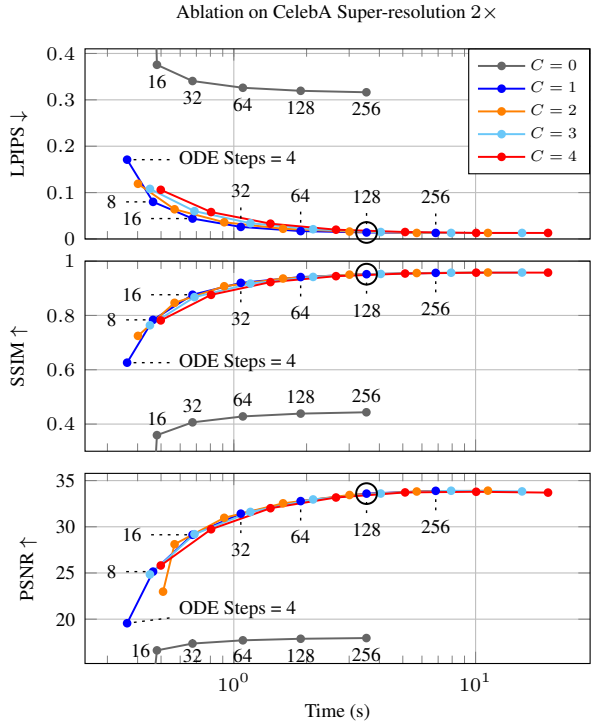}
    \caption{
    Ablation of ODE steps (indicated by markers) and correction steps $C$ for \( 2\times \) super-resolution on CelebA comparing LPIPS~$\downarrow$ (top), SSIM~$\uparrow$ (middle) and PSNR~$\uparrow$ (bottom) to processing time~$\downarrow$.
    ODE steps increase from left to right and represent 4, 8, 16, 32, 64, 128 and 256, respectively.
    For better visualization, ODE steps 4 and 8 when using $C=0$ are omitted. 
    The circle indicates the selected hyperparameters.
    Time is per image and displayed on a logarithmic scale.
    }
    \label{fig:ablation_celeba}
\end{figure}

\begin{figure*}[t]
    \centering
    \includegraphics[width=0.99\linewidth]{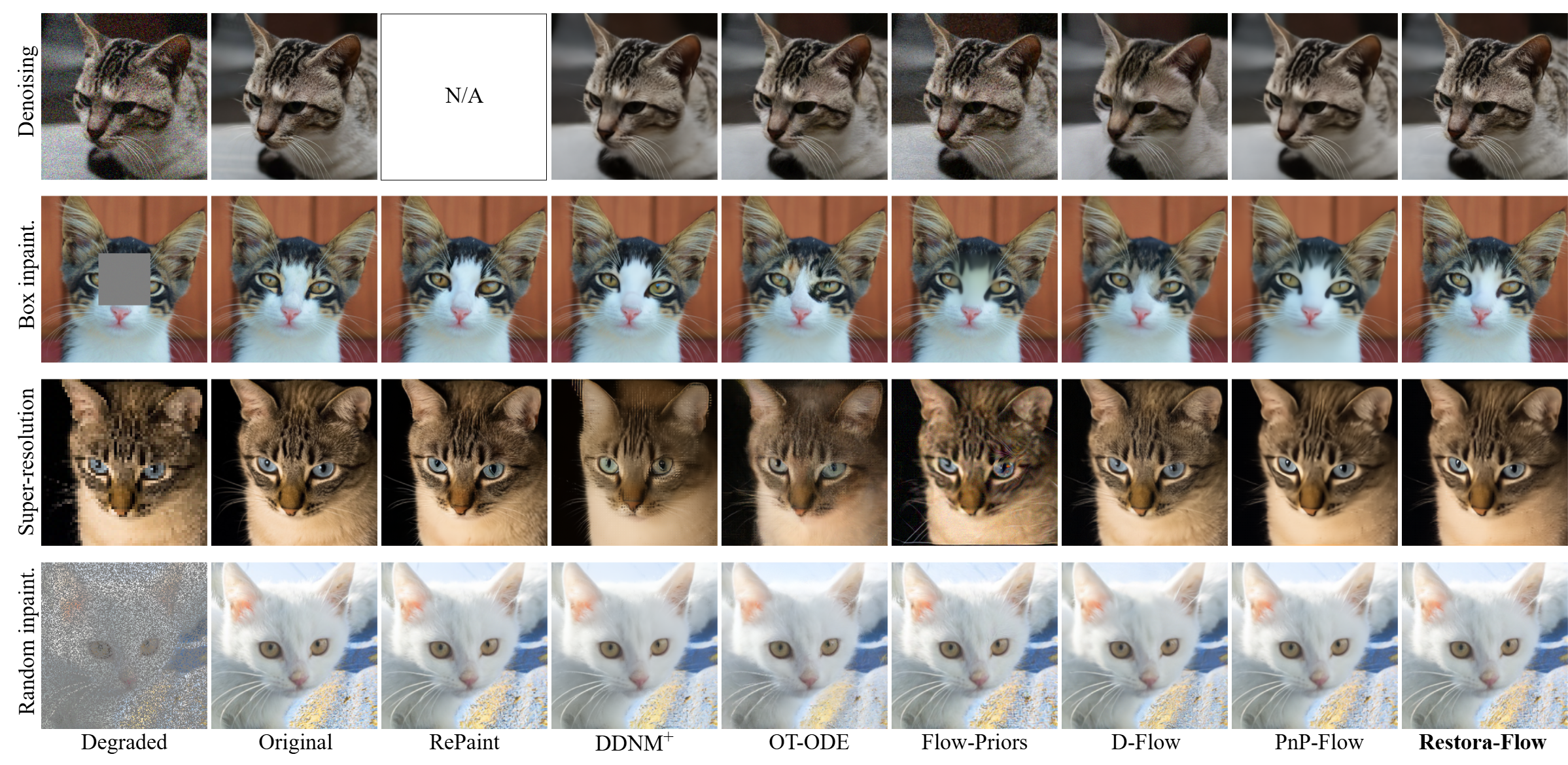}
    \caption{
    Qualitative results on AFHQ-Cat.
    Shown are the degraded image (col 1), original image (col 2), restored images using related work as indicated (cols 3-8) and restored images of Restora-Flow (col 9).
    Rows refer to denoising (row 1), box inpainting (row 2), super-resolution (row 3) and random inpainting (row 4).
    Differences can be best seen in the pdf version.
    }
    \label{fig:cat_results}
\end{figure*}

\paragraph{Implementation details.}
We utilize the pretrained flow-based CelebA and AFHQ-Cat models provided by~\cite{martin2024pnp}.
To enable a fair comparison with the diffusion-based RePaint~\cite{lugmayr2022repaint} and DDNM\textsuperscript{+}~\cite{wang2023zeroshot} methods, we train DDPMs from scratch using the same U-Net architecture and training parameters as the pretrained flow-based models. 
Specifically, for CelebA, we use a learning rate of \(1e\!-\!4\), a batch size of 128, and train for 200 epochs. 
For AFHQ-Cat, we use the same learning rate, a batch size of 64, and train for 400 epochs.
The number of diffusion time steps is set to 250 for both models. 
Additionally, we train a DDPM and a flow-based model from scratch on COCO, using a learning rate of \(1e\!-\!4\), a batch size of 64, and a total of 300 epochs.
We use the implementations provided by~\cite{martin2024pnp} for all flow-based reference methods. 
For a fair comparison, we integrate RePaint, DDNM\textsuperscript{+} and our proposed method, Restora-Flow, within this framework.
Regarding the medical X-ray Hand dataset, we use the pretrained flow-based and diffusion-based models from~\cite{hadzic2025flow} and implement all baselines and our proposed Restora-Flow within the MED-DDPM framework~\cite{dorjsembe2024three,hadzic2025flow}.

For our proposed method Restora-Flow, we use the same number of corrections (\ie, \(C= 1\), see \cref{fig:qualitative_correction_steps}), across all experiments and datasets, resulting in only one hyperparameter to optimize: the number of ODE steps (see our ablation in \cref{fig:ablation_celeba}). 
For all natural image datasets, we apply 64 ODE steps for denoising and box inpainting, 128 ODE steps for \(2\times\) super-resolution and random inpainting, and 256 ODE steps for \(4\times\) super-resolution. For the medical X-ray Hand dataset, we use 64 ODE steps for denoising and \(2\times\) super-resolution, and 32 ODE steps for box inpainting and occlusion removal. We trained generative prior models on NVIDIA A100 GPUs, which were also utilized for restoration experiments at $256\times256$ resolution. Experiments at $128\times128$ resolution were conducted using an NVIDIA GeForce RTX 3090.

\section{Results} 
\label{sec:results}
We evaluate the proposed Restora-Flow method by comparing it to related work: quantitatively on all datasets in \cref{tab:combined_results} and qualitatively on CelebA in \cref{fig:celeba_results} and AFHQ-Cat in \cref{fig:cat_results}. Qualitative results for COCO and X-ray Hand are provided in the supplemental material.
Plots visualizing reconstruction quality over processing time per metric of each considered method are shown for the CelebA dataset in \cref{fig:performance_vs_time_celeba}, for the AFHQ-Cat dataset in \cref{fig:overview} (left) and also in the supplemental material.
Ablations of hyperparameters are presented in \cref{fig:ablation_celeba}.

\paragraph{Quantitative Results.}
Our evaluation on the CelebA dataset in \cref{tab:combined_results} shows that Restora-Flow consistently outperforms all other flow-based methods in terms of perceptual quality, achieving the best LPIPS scores across all tasks. Moreover, our method achieves superior SSIM and PSNR distortion scores for denoising and super-resolution, while being a close second to Flow-Priors for box inpainting and to PnP-Flow for random inpainting. Further, it can be observed that diffusion-based RePaint delivers very competitive results in inpainting and super-resolution, while DDNM\textsuperscript{+} performs well on box inpainting. Nevertheless, our Restora-Flow method is roughly $10\times$ faster than RePaint and $6\times$ faster than DDNM\textsuperscript{+}, while delivering the same or even better reconstructions in terms of restoration quality. 
Additionally, when assessing quality scores and processing times simultaneously, the plots in \cref{fig:performance_vs_time_celeba} clearly illustrate the benefits of Restora-Flow compared with all other studied methods across all tasks.

The superior performance of Restora-Flow in comparison with related work can also be observed on the other two natural image datasets considered in our evaluation, which confirms the versatility of our approach. 
On the AFHQ-Cat and COCO datasets, which differ strongly in terms of variety of the images to reconstruct, Restora-Flow outperforms all other flow-based methods in the perceptual LPIPS metric for all tasks except for super-resolution on AFHQ-Cat and for box inpainting on COCO. 
However, in both cases our results are a close second to the respective best performing method. 
Moreover, when looking at the other metrics over these datasets, we also achieve the best results in all metrics and closely match the best performing method for PSNR on AFHQ-Cat. 
Importantly, these great results were achieved in tandem with the fastest processing times in most settings. OT-ODE is the only related work that was able to be faster in two settings on AFHQ-Cat, however, it achieved inferior reconstruction quality in both cases. 
For diffusion-based baselines, we observe similar behavior as for CelebA, often delivering competitive results but at a significantly higher computational cost. 
The relationship between perceptual quality scores and processing times for AFHQ-Cat is visualized in \cref{fig:overview}, further demonstrating the advantages of our method compared to related work.
For additional plots on AFHQ-Cat using distortion metrics, see suppl. material.

To further assess the versatility of our method, and to additionally study the dependence of the Restora-Flow algorithm on the specific implementation framework, we performed an experiment on a medical X-ray Hand dataset and integrated all baselines and our proposed method in the Med-DDPM framework (see \cref{sec:experimental_setup}). 
In-line with previously discussed results, our method consistently achieved great results for LPIPS, SSIM and PSNR compared with the baselines in a fraction of the time, see \cref{tab:combined_results}.
The same behavior was also observed in the clinically motivated occlusion removal task, where downstream segmentation or classification tasks may benefit from imaging artifacts or anomalies being removed.
Consistent with the other experiments, Restora-Flow achieved improved perceptual quality scores at lower processing times.

\paragraph{Ablation.}
\cref{fig:ablation_celeba} investigates the influence of ODE steps and correction steps $C$ for \( 2\times \) super-resolution on CelebA shown for LPIPS (top) and PSNR (bottom).
ODE steps (indicated by markers) increase from left to right from 4 up to 256 for each $C$.
As expected, more correction steps lead to longer evaluation times using the same number of ODE steps.
Both graphs show a trade-off between fast sampling and better scores, with lower correction steps being advantageous despite requiring a larger number of ODE steps until the metrics saturate.
Therefore, by setting $C=1$ for all experiments, the only hyperparameter that remains is the number of ODE steps, as in standard flow matching.

\paragraph{Qualitative Results.}
The qualitative results in \cref{fig:celeba_results}, \cref{fig:cat_results} and Suppl. Fig. 1 indicate notable differences in performance across different tasks and datasets among the studied baseline methods. 
OT-ODE tends to produce artifacts, particularly evident in box inpainting outputs on the CelebA and AFHQ-Cat datasets, as well as in the random inpainting task on the COCO dataset. Similarly, Flow-Priors generates noisy reconstructions in super-resolution outputs on the CelebA dataset, and produces artifacts in both super-resolution and random inpainting tasks on the AFHQ-Cat and COCO datasets, respectively. While D-Flow generally yields realistic results, it suffers from slow processing times and struggles to reconstruct certain objects for denoising on the COCO dataset. Despite achieving high SSIM and PSNR scores, PnP-Flow often produces over-smoothed results across all datasets.
Among the diffusion-based baselines, DDNM\textsuperscript{+} tends to introduce artifacts in super-resolution tasks, while RePaint produces visually realistic results but at the expense of very slow processing.
In contrast, our method was able to generate artifact-free and realistic images that maintain texture while ensuring fast processing across conducted experiments (see also \cref{fig:overview}).
Additional examples are in the supplemental material.

For the medical dataset, (see \cref{fig:overview} and Suppl. Fig. 3), we observe lower overall variability among methods, likely due to the smaller variations present in hand X-ray images compared with natural images. 
Nonetheless, notable differences, which may be clinically significant due to anatomical changes, are evident in some tasks.
For example, although D-Flow reconstructions initially appear realistic, the known input regions are often altered substantially (\eg, in box inpainting), likely due to sensitive hyperparameters. In addition, several baselines struggle in the occlusion removal task, where occlusions remain partially visible in the reconstructed images. In contrast, our method consistently produces high-quality reconstructions across all tasks, with considerably faster processing times.

\section{Conclusion}
\label{sec:conclusion}
We have introduced Restora-Flow, a flow matching-based algorithm designed for mask-guided image restoration.
By effectively incorporating a trajectory correction mechanism within the mask-guided generation of a flow prior, our method has shown superior perceptual quality and reduced processing time across various tasks and datasets when compared to diffusion and flow-based baselines. 
Moreover, Restora-Flow requires no additional hyperparameters beyond the number of ODE steps utilized in standard flow matching.
In future work, we plan to extend the algorithm to tackle image restoration tasks that involve non mask-based degradation operators.

\section*{Acknowledgements}
We would like to thank Mateusz Kozinski for constructive discussions and feedback that contributed to this work. 
This research was funded in whole or in part by the Austrian Science Fund (FWF) 10.55776/PAT1748423. 

{
    \small
    \bibliographystyle{ieeenat_fullname}
    \bibliography{main}
}
\renewcommand{\figurename}{Supplemental Figure}
\onecolumn

\clearpage
\section*{Appendix}

\vspace{1cm}
\begin{figure*}[h]
    \centering
    \includegraphics[width=0.99\linewidth]{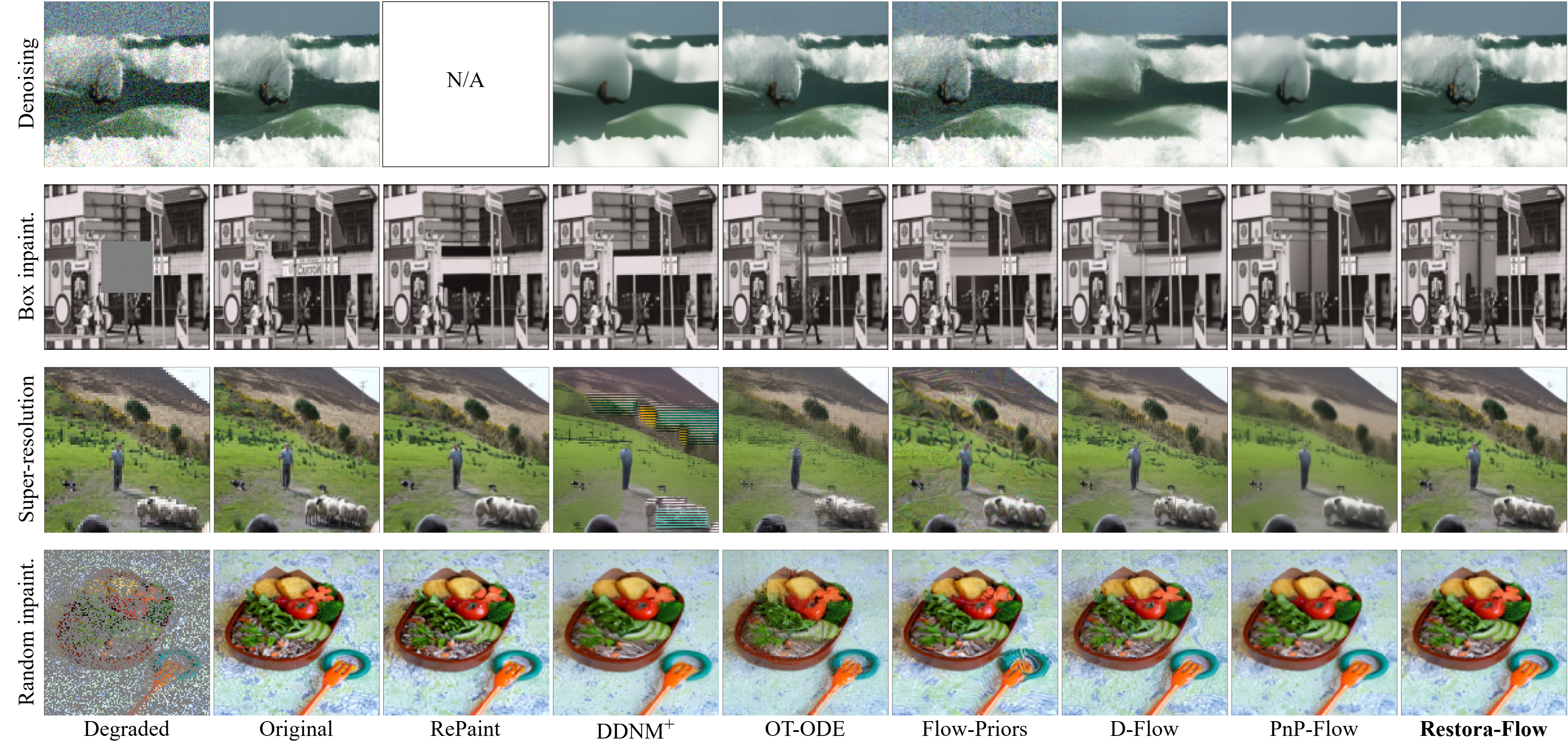}
    \captionof{figure}{
    Qualitative results on COCO.
    Shown are the degraded image (col 1), original image (col 2), restored images using related work as indicated (cols 3-8) and restored images of Restora-Flow (col 9).
    Rows refer to denoising (row 1), box inpainting (row 2), super-resolution (row 3) and random inpainting (row 4).
    Differences can be best seen in the pdf version.
    }
    \label{fig:coco_results}
\end{figure*}

\vspace{1cm}
\begin{figure*}[h]
    \begin{minipage}{.32\textwidth}
        \centering
        \definecolor{repaint}{HTML}{955196}
\definecolor{otode}{HTML}{87A96B}
\definecolor{flowpriors}{HTML}{CFB53B}
\definecolor{dflow}{HTML}{FF9966}
\definecolor{pnpflow}{HTML}{6699CC}
\definecolor{restoraflow}{HTML}{CC3333}

\definecolor{denoising}{HTML}{9466FF}
\definecolor{boxinpainting}{HTML}{69C7F8}
\definecolor{superresolution}{HTML}{D93F50}
\definecolor{randominpainting}{HTML}{FF9966}

\def\myopacity{1}

\begin{tikzpicture}[scale=0.59]
    \begin{groupplot}[
        group style={
            group name=group,
            group size=2 by 1,
            horizontal sep=10pt
        },
        ymin=0, ymax=0.5,
        yticklabel style={/pgf/number format/fixed, /pgf/number format/precision=5},
        height=7cm,
        grid=major,
    ]

    \nextgroupplot[xmax=50,
        xtick={0, 10, 20, 30, 40, 50},
        xticklabels={0, 10, 20, 30, 40},
        ytick={0, 0.1, 0.2, 0.3, 0.4, 0.5},
        axis y line*=left,
        width=6cm,
        mark options={scale=1.5}
    ]

    \addplot[only marks, mark=+, mark size=1.75, color=denoising, opacity=\myopacity] coordinates {(13.74, 0.170)};
    \addplot[only marks, mark=*, mark size=1.75, color=denoising] coordinates {(5.54, 0.078)};
    \addplot[only marks, mark=square*, mark size=1.5, color=denoising] coordinates {(67.10, 0.153)};
    \addplot[only marks, mark=diamond*, color=denoising] coordinates {(44.45, 0.184)};
    \addplot[only marks, mark=triangle*, color=denoising] coordinates {(9.86, 0.165)};
    \addplot[only marks, mark=pentagon*, mark options={draw=black, line width=0.05mm, fill=denoising, opacity=\myopacity}, mark size=4] coordinates {(0.72, 0.051)};

    \addplot[only marks, mark=asterisk, color=boxinpainting, opacity=\myopacity] coordinates {(86.23, 0.043)};
    \addplot[only marks, mark=+, mark size=1.75, color=boxinpainting, opacity=\myopacity] coordinates {(13.74, 0.048)};
    \addplot[only marks, mark=*, mark size=1.75, color=boxinpainting] coordinates {(6.94, 0.048)};
    \addplot[only marks, mark=square*, mark size=1.5, color=boxinpainting] coordinates {(67.05, 0.054)};
    \addplot[only marks, mark=diamond*, color=boxinpainting] coordinates {(126.09, 0.112)};
    \addplot[only marks, mark=triangle*, color=boxinpainting] coordinates {(9.86, 0.124)};
    \addplot[only marks, mark=pentagon*, mark options={draw=black, line width=0.05mm, fill=boxinpainting, opacity=\myopacity}, mark size=4] coordinates {(3.96, 0.047)};

    \addplot[only marks, mark=asterisk, color=superresolution, opacity=\myopacity] coordinates {(86.23, 0.139)};
    \addplot[only marks, mark=+, mark size=1.75, color=superresolution, opacity=\myopacity] coordinates {(13.74, 0.462)};
    \addplot[only marks, mark=*, mark size=1.75, color=superresolution] coordinates {(7.28, 0.285)};
    \addplot[only marks, mark=square*, mark size=1.5, color=superresolution] coordinates {(67.30, 0.271)};
    \addplot[only marks, mark=diamond*, color=superresolution] coordinates {(261.84, 0.123)};
    \addplot[only marks, mark=triangle*, color=superresolution] coordinates {(46.26, 0.180)};
    \addplot[only marks, mark=pentagon*, mark options={draw=black, line width=0.05mm, fill=superresolution, opacity=\myopacity}, mark size=4] coordinates {(14.48, 0.158)};

    \addplot[only marks, mark=asterisk, color=randominpainting, opacity=\myopacity] coordinates {(86.23, 0.034)};
    \addplot[only marks, mark=+, mark size=1.75, color=randominpainting, opacity=\myopacity] coordinates {(13.74, 0.065)};
    \addplot[only marks, mark=*, mark size=1.75, color=randominpainting] coordinates {(7.28, 0.094)};
    \addplot[only marks, mark=square*, mark size=1.5, color=randominpainting] coordinates {(67.69, 0.046)};
    \addplot[only marks, mark=diamond*, color=randominpainting] coordinates {(266.18, 0.056)};
    \addplot[only marks, mark=triangle*, color=randominpainting] coordinates {(19.15, 0.042)};
    \addplot[only marks, mark=pentagon*, mark options={draw=black, line width=0.05mm, fill=randominpainting, opacity=\myopacity}, mark size=4] coordinates {(7.48, 0.034)};

    \nextgroupplot[xmin=50,xmax=300,
        xtick={50, 100, 200, 300},
        ylabel={},
        ytick={},
        yticklabels={},
        axis y line*=right,
        width=4cm,
        mark options={scale=1.5}
    ]

    \addplot[only marks, mark=+, mark size=1.75, color=denoising, opacity=\myopacity] coordinates {(13.74, 0.170)};
    \addplot[only marks, mark=*, mark size=1.75, color=denoising] coordinates {(5.54, 0.078)};
    \addplot[only marks, mark=square*, mark size=1.5, color=denoising] coordinates {(67.10, 0.153)};
    \addplot[only marks, mark=diamond*, color=denoising] coordinates {(44.45, 0.184)};
    \addplot[only marks, mark=triangle*, color=denoising] coordinates {(9.86, 0.165)};
    \addplot[only marks, mark=pentagon*, mark options={draw=black, line width=0.05mm, fill=denoising, opacity=\myopacity}, mark size=4] coordinates {(0.72, 0.051)};

    \addplot[only marks, mark=asterisk, color=boxinpainting, opacity=\myopacity] coordinates {(86.23, 0.043)};
    \addplot[only marks, mark=+, mark size=1.75, color=boxinpainting, opacity=\myopacity] coordinates {(13.74, 0.048)};
    \addplot[only marks, mark=*, mark size=1.75, color=boxinpainting] coordinates {(6.94, 0.048)};
    \addplot[only marks, mark=square*, mark size=1.5, color=boxinpainting] coordinates {(67.05, 0.054)};
    \addplot[only marks, mark=diamond*, color=boxinpainting] coordinates {(126.09, 0.112)};
    \addplot[only marks, mark=triangle*, color=boxinpainting] coordinates {(9.86, 0.124)};
    \addplot[only marks, mark=pentagon*, mark options={draw=black, line width=0.05mm, fill=boxinpainting, opacity=\myopacity}, mark size=4] coordinates {(3.96, 0.047)};

    \addplot[only marks, mark=asterisk, color=superresolution, opacity=\myopacity] coordinates {(86.23, 0.139)};
    \addplot[only marks, mark=+, mark size=1.75, color=superresolution, opacity=\myopacity] coordinates {(13.74, 0.462)};
    \addplot[only marks, mark=*, mark size=1.75, color=superresolution] coordinates {(7.28, 0.285)};
    \addplot[only marks, mark=square*, mark size=1.5, color=superresolution] coordinates {(67.30, 0.271)};
    \addplot[only marks, mark=diamond*, color=superresolution] coordinates {(261.84, 0.123)};
    \addplot[only marks, mark=triangle*, color=superresolution] coordinates {(46.26, 0.180)};
    \addplot[only marks, mark=pentagon*, mark options={draw=black, line width=0.05mm, fill=superresolution, opacity=\myopacity}, mark size=4] coordinates {(14.48, 0.158)};

    \addplot[only marks, mark=asterisk, color=randominpainting, opacity=\myopacity] coordinates {(86.23, 0.034)};
    \addplot[only marks, mark=+, mark size=1.75, color=randominpainting, opacity=\myopacity] coordinates {(13.74, 0.065)};
    \addplot[only marks, mark=*, mark size=1.75, color=randominpainting] coordinates {(7.28, 0.094)};
    \addplot[only marks, mark=square*, mark size=1.5, color=randominpainting] coordinates {(67.69, 0.046)};
    \addplot[only marks, mark=diamond*, color=randominpainting] coordinates {(266.18, 0.056)};
    \addplot[only marks, mark=triangle*, color=randominpainting] coordinates {(19.15, 0.042)};
    \addplot[only marks, mark=pentagon*, mark options={draw=black, line width=0.05mm, fill=randominpainting, opacity=\myopacity}, mark size=4] coordinates {(7.48, 0.034)};

    \end{groupplot}

    \node (title) at ($(group c1r1.north)!0.5!(group c2r1.north)+(-0.8,0.5cm)$) {\scriptsize AFHQ-Cat: LPIPS $\downarrow$};
    \node (xlabel) at ($(group c1r1.south)!0.5!(group c2r1.south)+(-0.8,-0.8cm)$) {\scriptsize Time (s)};
    \node[rotate=90] (ylabel) at ($(group c1r1.west)+(-1.2cm,0)$) {\scriptsize Score};

    \node[inner sep=0, anchor=north west] (tilde) at ($(group c1r1.east)+(-0.01cm,0)$) {\tiny $\sim$};

\end{tikzpicture}
    \end{minipage}
    \hspace{2pt}
    \begin{minipage}{.32\textwidth}
        \centering
        \definecolor{repaint}{HTML}{955196}
\definecolor{otode}{HTML}{87A96B}
\definecolor{flowpriors}{HTML}{CFB53B}
\definecolor{dflow}{HTML}{FF9966}
\definecolor{pnpflow}{HTML}{6699CC}
\definecolor{restoraflow}{HTML}{CC3333}

\definecolor{denoising}{HTML}{9466FF}
\definecolor{boxinpainting}{HTML}{69C7F8}
\definecolor{superresolution}{HTML}{D93F50}
\definecolor{randominpainting}{HTML}{FF9966}

\def\myopacity{1}

\begin{tikzpicture}[scale=0.59]
    \begin{groupplot}[
        group style={
            group name=group,
            group size=2 by 1,
            horizontal sep=10pt
        },
        ymin=0.5, ymax=1,
        yticklabel style={/pgf/number format/fixed, /pgf/number format/precision=5},
        height=7cm,
        grid=major,
    ]

    \nextgroupplot[xmax=50,
        xtick={0, 10, 20, 30, 40, 50},
        xticklabels={0, 10, 20, 30, 40},
        ytick={0.5, 0.6, 0.7, 0.8, 0.9, 1},
        axis y line*=left,
        width=6cm,
        mark options={scale=1.5}
    ]

    \addplot[only marks, mark=+, mark size=1.75, color=denoising] coordinates {(13.74, 0.818)};
    \addplot[only marks, mark=*, mark size=1.75, color=denoising] coordinates {(5.54, 0.814)};
    \addplot[only marks, mark=square*, mark size=1.5, color=denoising] coordinates {(67.10, 0.771)};
    \addplot[only marks, mark=diamond*, color=denoising] coordinates {(44.45, 0.648)};
    \addplot[only marks, mark=triangle*, color=denoising] coordinates {(9.86, 0.864)};
    \addplot[only marks, mark=pentagon*, mark options={draw=black, line width=0.05mm, fill=denoising, opacity=\myopacity}, mark size=4] coordinates {(0.72, 0.899)};

    \addplot[only marks, mark=asterisk, color=boxinpainting, opacity=\myopacity] coordinates {(86.23, 0.939)};
    \addplot[only marks, mark=+, mark size=1.75, color=boxinpainting] coordinates {(13.74, 0.942)};
    \addplot[only marks, mark=*, mark size=1.75, color=boxinpainting] coordinates {(6.94, 0.924)};
    \addplot[only marks, mark=square*, mark size=1.5, color=boxinpainting] coordinates {(67.05, 0.942)};
    \addplot[only marks, mark=diamond*, color=boxinpainting] coordinates {(126.09, 0.839)};
    \addplot[only marks, mark=triangle*, color=boxinpainting] coordinates {(9.86, 0.904)};
    \addplot[only marks, mark=pentagon*, mark options={draw=black, line width=0.05mm, fill=boxinpainting, opacity=\myopacity}, mark size=4] coordinates {(3.96, 0.939)};

    \addplot[only marks, mark=asterisk, color=superresolution, opacity=\myopacity] coordinates {(86.23, 0.701)};
    \addplot[only marks, mark=+, mark size=1.75, color=superresolution] coordinates {(13.74, 0.534)};
    \addplot[only marks, mark=*, mark size=1.75, color=superresolution] coordinates {(7.28, 0.565)};
    \addplot[only marks, mark=square*, mark size=1.5, color=superresolution] coordinates {(67.30, 0.565)};
    \addplot[only marks, mark=diamond*, color=superresolution] coordinates {(261.84, 0.707)};
    \addplot[only marks, mark=triangle*, color=superresolution] coordinates {(46.26, 0.790)};
    \addplot[only marks, mark=pentagon*, mark options={draw=black, line width=0.05mm, fill=superresolution, opacity=\myopacity}, mark size=4] coordinates {(14.48, 0.761)};

    \addplot[only marks, mark=asterisk, color=randominpainting, opacity=\myopacity] coordinates {(86.23, 0.897)};
    \addplot[only marks, mark=+, mark size=1.75, color=randominpainting] coordinates {(13.74, 0.876)};
    \addplot[only marks, mark=*, mark size=1.75, color=randominpainting] coordinates {(7.28, 0.839)};
    \addplot[only marks, mark=square*, mark size=1.5, color=randominpainting] coordinates {(67.69, 0.909)};
    \addplot[only marks, mark=diamond*, color=randominpainting] coordinates {(266.18, 0.878)};
    \addplot[only marks, mark=triangle*, color=randominpainting] coordinates {(19.15, 0.930)};
    \addplot[only marks, mark=pentagon*, mark options={draw=black, line width=0.05mm, fill=randominpainting, opacity=\myopacity}, mark size=4] coordinates {(7.48, 0.914)};
    
    \nextgroupplot[xmin=50,xmax=300,
        xtick={50, 100, 200, 300},
        ylabel={},
        ytick={},
        yticklabels={},
        axis y line*=right,
        width=4cm,
        mark options={scale=1.5}
    ]

    \addplot[only marks, mark=+, mark size=1.75, color=denoising] coordinates {(13.74, 0.818)};
    \addplot[only marks, mark=*, mark size=1.75, color=denoising] coordinates {(5.54, 0.814)};
    \addplot[only marks, mark=square*, mark size=1.5, color=denoising] coordinates {(67.10, 0.771)};
    \addplot[only marks, mark=diamond*, color=denoising] coordinates {(44.45, 0.648)};
    \addplot[only marks, mark=triangle*, color=denoising] coordinates {(9.86, 0.864)};
    \addplot[only marks, mark=pentagon*, mark options={draw=black, line width=0.05mm, fill=denoising, opacity=\myopacity}, mark size=4] coordinates {(0.72, 0.899)};

    \addplot[only marks, mark=asterisk, color=boxinpainting, opacity=\myopacity] coordinates {(86.23, 0.939)};
    \addplot[only marks, mark=+, mark size=1.75, color=boxinpainting] coordinates {(13.74, 0.942)};
    \addplot[only marks, mark=*, mark size=1.75, color=boxinpainting] coordinates {(6.94, 0.924)};
    \addplot[only marks, mark=square*, mark size=1.5, color=boxinpainting] coordinates {(67.05, 0.942)};
    \addplot[only marks, mark=diamond*, color=boxinpainting] coordinates {(126.09, 0.839)};
    \addplot[only marks, mark=triangle*, color=boxinpainting] coordinates {(9.86, 0.904)};
    \addplot[only marks, mark=pentagon*, mark options={draw=black, line width=0.05mm, fill=boxinpainting, opacity=\myopacity}, mark size=4] coordinates {(3.96, 0.939)};

    \addplot[only marks, mark=asterisk, color=superresolution, opacity=\myopacity] coordinates {(86.23, 0.701)};
    \addplot[only marks, mark=+, mark size=1.75, color=superresolution] coordinates {(13.74, 0.534)};
    \addplot[only marks, mark=*, mark size=1.75, color=superresolution] coordinates {(7.28, 0.565)};
    \addplot[only marks, mark=square*, mark size=1.5, color=superresolution] coordinates {(67.30, 0.565)};
    \addplot[only marks, mark=diamond*, color=superresolution] coordinates {(261.84, 0.707)};
    \addplot[only marks, mark=triangle*, color=superresolution] coordinates {(46.26, 0.790)};
    \addplot[only marks, mark=pentagon*, mark options={draw=black, line width=0.05mm, fill=superresolution, opacity=\myopacity}, mark size=4] coordinates {(14.48, 0.761)};

    \addplot[only marks, mark=asterisk, color=randominpainting, opacity=\myopacity] coordinates {(86.23, 0.897)};
    \addplot[only marks, mark=+, mark size=1.75, color=randominpainting] coordinates {(13.74, 0.876)};
    \addplot[only marks, mark=*, mark size=1.75, color=randominpainting] coordinates {(7.28, 0.839)};
    \addplot[only marks, mark=square*, mark size=1.5, color=randominpainting] coordinates {(67.69, 0.909)};
    \addplot[only marks, mark=diamond*, color=randominpainting] coordinates {(266.18, 0.878)};
    \addplot[only marks, mark=triangle*, color=randominpainting] coordinates {(19.15, 0.930)};
    \addplot[only marks, mark=pentagon*, mark options={draw=black, line width=0.05mm, fill=randominpainting, opacity=\myopacity}, mark size=4] coordinates {(7.48, 0.914)};

    \end{groupplot}

    \node (title) at ($(group c1r1.north)!0.5!(group c2r1.north)+(-0.8,0.5cm)$) {\scriptsize AFHQ-Cat: SSIM $\uparrow$};
    \node (xlabel) at ($(group c1r1.south)!0.5!(group c2r1.south)+(-0.8,-0.8cm)$) {\scriptsize Time (s)};
    \node[rotate=90] (ylabel) at ($(group c1r1.west)+(-1.2cm,0)$) {\scriptsize Score};

    \node[inner sep=0, anchor=north west] (tilde) at ($(group c1r1.east)+(-0.01cm,0)$) {\tiny $\sim$};
    
\end{tikzpicture}
    \end{minipage}
    \hspace{2pt}
    \begin{minipage}{.32\textwidth}
        \centering
        \definecolor{repaint}{HTML}{955196}
\definecolor{otode}{HTML}{87A96B}
\definecolor{flowpriors}{HTML}{CFB53B}
\definecolor{dflow}{HTML}{FF9966}
\definecolor{pnpflow}{HTML}{6699CC}
\definecolor{restoraflow}{HTML}{CC3333}

\definecolor{denoising}{HTML}{9466FF}
\definecolor{boxinpainting}{HTML}{69C7F8}
\definecolor{superresolution}{HTML}{D93F50}
\definecolor{randominpainting}{HTML}{FF9966}

\def\myopacity{1}

\begin{tikzpicture}[scale=0.59]
    \begin{groupplot}[
        group style={
            group name=group,
            group size=2 by 1,
            horizontal sep=10pt
        },
        ymin=18, ymax=35,
        yticklabel style={/pgf/number format/fixed, /pgf/number format/precision=5},
        height=7cm,
        grid=major,
    ]

    \nextgroupplot[xmax=50,
        xtick={0, 10, 20, 30, 40, 50},
        xticklabels={0, 10, 20, 30, 40},
        ytick={20, 25, 30, 35},
        axis y line*=left,
        width=6cm,
        mark options={scale=1.5},
        legend style={at={(2.05, .02)}, anchor=south east},
        legend cell align={left},
    ]



    \addplot[only marks, mark=+, mark size=1.75, color=denoising, opacity=\myopacity] coordinates {(13.74, 29.06)};
    
    \addplot[only marks, mark=*, mark size=1.75, color=denoising] coordinates {(5.54, 29.73)};

    \addplot[only marks, mark=square*, mark size=1.5, color=denoising] coordinates {(67.10, 29.43)};
    \addplot[only marks, mark=diamond*, color=denoising] coordinates {(44.45, 24.98)};
    \addplot[only marks, mark=triangle*, color=denoising] coordinates {(9.86, 31.10)};
    \addplot[only marks, mark=pentagon*, mark options={draw=black, line width=0.05mm, fill=denoising, opacity=\myopacity}, mark size=4] coordinates {(0.72, 32.25)};

    \addplot[only marks, mark=asterisk, color=boxinpainting, opacity=\myopacity] coordinates {(86.23, 26.26)};
    \addplot[only marks, mark=+, mark size=1.75, color=boxinpainting, opacity=\myopacity] coordinates {(13.74, 25.16)};
    \addplot[only marks, mark=*, mark size=1.75, color=boxinpainting] coordinates {(6.94, 24.36)};
    \addplot[only marks, mark=square*, mark size=1.5, color=boxinpainting] coordinates {(67.05, 26.04)};
    \addplot[only marks, mark=diamond*, color=boxinpainting] coordinates {(126.09, 26.17)};
    \addplot[only marks, mark=triangle*, color=boxinpainting] coordinates {(9.86, 26.18)};
    \addplot[only marks, mark=pentagon*, mark options={draw=black, line width=0.05mm, fill=boxinpainting, opacity=\myopacity}, mark size=4] coordinates {(3.96, 25.96)};

    \addplot[only marks, mark=asterisk, color=superresolution, opacity=\myopacity] coordinates {(86.23, 24.71)};
    \addplot[only marks, mark=+, mark size=1.75, color=superresolution, opacity=\myopacity] coordinates {(13.74, 19.69)};
    \addplot[only marks, mark=*, mark size=1.75, color=superresolution] coordinates {(7.28, 21.85)};
    \addplot[only marks, mark=square*, mark size=1.5, color=superresolution] coordinates {(67.30, 23.50)};
    \addplot[only marks, mark=diamond*, color=superresolution] coordinates {(281.84, 25.34)};
    \addplot[only marks, mark=triangle*, color=superresolution] coordinates {(46.26, 26.95)};
    \addplot[only marks, mark=pentagon*, mark options={draw=black, line width=0.05mm, fill=superresolution, opacity=\myopacity}, mark size=4] coordinates {(14.48, 26.33)};

    \addplot[only marks, mark=asterisk, color=randominpainting, opacity=\myopacity] coordinates {(86.23, 30.93)};
    \addplot[only marks, mark=+, mark size=1.75, color=randominpainting, opacity=\myopacity] coordinates {(13.74, 30.12)};
    \addplot[only marks, mark=*, mark size=1.75, color=randominpainting] coordinates {(7.28, 28.87)};
    \addplot[only marks, mark=square*, mark size=1.5, color=randominpainting] coordinates {(67.69, 31.82)};
    \addplot[only marks, mark=diamond*, color=randominpainting] coordinates {(266.18, 30.97)};
    \addplot[only marks, mark=triangle*, color=randominpainting] coordinates {(19.15, 33.07)};
    \addplot[only marks, mark=pentagon*, mark options={draw=black, line width=0.05mm, fill=randominpainting, opacity=\myopacity}, mark size=4] coordinates {(7.48, 31.99)};
    
    \nextgroupplot[xmin=50,xmax=300,
        xtick={50, 100, 200, 300},
        ylabel={},
        ytick={},
        yticklabels={},
        axis y line*=right,
        width=4cm,
        mark options={scale=1.5}
    ]

    \addplot[only marks, mark=+, mark size=1.75, color=denoising, opacity=\myopacity] coordinates {(13.74, 29.06)};
    
    \addplot[only marks, mark=*, mark size=1.75, color=denoising] coordinates {(5.54, 29.73)};

    \addplot[only marks, mark=square*, mark size=1.5, color=denoising] coordinates {(67.10, 29.43)};
    \addplot[only marks, mark=diamond*, color=denoising] coordinates {(44.45, 24.98)};
    \addplot[only marks, mark=triangle*, color=denoising] coordinates {(9.86, 31.10)};
    \addplot[only marks, mark=pentagon*, mark options={draw=black, line width=0.05mm, fill=denoising, opacity=\myopacity}, mark size=4] coordinates {(0.72, 32.25)};

    \addplot[only marks, mark=asterisk, color=boxinpainting, opacity=\myopacity] coordinates {(86.23, 26.26)};
    \addplot[only marks, mark=+, mark size=1.75, color=boxinpainting, opacity=\myopacity] coordinates {(13.74, 25.16)};
    \addplot[only marks, mark=*, mark size=1.75, color=boxinpainting] coordinates {(6.94, 24.36)};
    \addplot[only marks, mark=square*, mark size=1.5, color=boxinpainting] coordinates {(67.05, 26.04)};
    \addplot[only marks, mark=diamond*, color=boxinpainting] coordinates {(126.09, 26.17)};
    \addplot[only marks, mark=triangle*, color=boxinpainting] coordinates {(9.86, 26.18)};
    \addplot[only marks, mark=pentagon*, mark options={draw=black, line width=0.05mm, fill=boxinpainting, opacity=\myopacity}, mark size=4] coordinates {(3.96, 25.96)};

    \addplot[only marks, mark=asterisk, color=superresolution, opacity=\myopacity] coordinates {(86.23, 24.71)};
    \addplot[only marks, mark=+, mark size=1.75, color=superresolution, opacity=\myopacity] coordinates {(13.74, 20.01)};   
    \addplot[only marks, mark=*, mark size=1.75, color=superresolution] coordinates {(7.28, 21.85)};
    \addplot[only marks, mark=square*, mark size=1.5, color=superresolution] coordinates {(67.30, 23.50)};
    \addplot[only marks, mark=diamond*, color=superresolution] coordinates {(281.84, 25.34)};
    \addplot[only marks, mark=triangle*, color=superresolution] coordinates {(46.26, 26.95)};
    \addplot[only marks, mark=pentagon*, mark options={draw=black, line width=0.05mm, fill=superresolution, opacity=\myopacity}, mark size=4] coordinates {(14.48, 26.33)};

    \addplot[only marks, mark=asterisk, color=randominpainting, opacity=\myopacity] coordinates {(86.23, 30.93)};
    \addplot[only marks, mark=+, mark size=1.75, color=randominpainting, opacity=\myopacity] coordinates {(13.74, 30.12)};
    \addplot[only marks, mark=*, mark size=1.75, color=randominpainting] coordinates {(7.28, 28.87)};
    \addplot[only marks, mark=square*, mark size=1.5, color=randominpainting] coordinates {(67.69, 31.82)};
    \addplot[only marks, mark=diamond*, color=randominpainting] coordinates {(266.18, 30.97)};
    \addplot[only marks, mark=triangle*, color=randominpainting] coordinates {(19.15, 33.07)};
    \addplot[only marks, mark=pentagon*, mark options={draw=black, line width=0.05mm, fill=randominpainting, opacity=\myopacity}, mark size=4] coordinates {(7.48, 31.99)};

    \end{groupplot}

    \node (title) at ($(group c1r1.north)!0.5!(group c2r1.north)+(-0.8,0.5cm)$) {\scriptsize AFHQ-Cat: PSNR $\uparrow$};
    \node (xlabel) at ($(group c1r1.south)!0.5!(group c2r1.south)+(-0.8,-0.8cm)$) {\scriptsize Time (s)};
    \node[rotate=90] (ylabel) at ($(group c1r1.west)+(-1.2cm,0)$) {\scriptsize Score};

    \node[inner sep=0, anchor=north west] (tilde) at ($(group c1r1.east)+(-0.01cm,0)$) {\tiny $\sim$};
    
\end{tikzpicture}
    \end{minipage}

        \centering
        \definecolor{repaint}{HTML}{955196}
\definecolor{otode}{HTML}{87A96B}
\definecolor{flowpriors}{HTML}{CFB53B}
\definecolor{dflow}{HTML}{FF9966}
\definecolor{pnpflow}{HTML}{6699CC}
\definecolor{restoraflow}{HTML}{CC3333}

\definecolor{denoising}{HTML}{9466FF}
\definecolor{boxinpainting}{HTML}{69C7F8}
\definecolor{superresolution}{HTML}{D93F50}
\definecolor{randominpainting}{HTML}{FF9966}

    



\begin{tikzpicture}[scale=0.63]
    \begin{axis}[%
    legend columns=6, 
    hide axis,
    xmin=10,
    xmax=100,
    ymin=0,
    ymax=0.4,
    height=2cm,
    width=10cm,
    legend style={draw=white!15!black, legend cell align=left}
    ]
    

    \addlegendimage{mark=asterisk,only marks,mark size=2}
    \addlegendentry{RePaint \phantom{x}}
    \addlegendimage{mark=square,only marks,mark size=1.75}
    \addlegendentry{Flow-Priors \phantom{x}}
    \addlegendimage{mark=triangle,only marks,mark size=2.5}
    \addlegendentry{PnP-Flow \phantom{x}}
    \addlegendimage{mark=pentagon,only marks,mark size=2.5}
    \addlegendentry{\textbf{Restora-Flow} \phantom{x}}
    \addlegendimage{empty legend}
    \addlegendentry[color=denoising]{Denoising \phantom{x}}
    \addlegendimage{empty legend}
    \addlegendentry[color=superresolution]{Super-resolution \phantom{x}}
    \addlegendimage{mark=+,only marks,mark size=1.75}
    \addlegendentry{DDNM\textsuperscript{+} \phantom{x}}
    \addlegendimage{mark=o,only marks,mark size=2}
    \addlegendentry{OT-ODE \phantom{x}}
    \addlegendimage{mark=diamond,only marks,mark size=2.5}
    \addlegendentry{D-Flow \phantom{x}}
    \addlegendimage{empty legend}
    \addlegendentry[color=boxinpainting]{\phantom{x}}
    \addlegendimage{empty legend}
    \addlegendentry[color=boxinpainting]{Box inpainting \phantom{x}}
    \addlegendimage{empty legend}
    \addlegendentry[color=randominpainting]{Random inpainting}
    
    \end{axis}
\end{tikzpicture}
    \caption{
    Visual representation of quantitative results on CelebA.
    Restora-Flow (\pentagon) is compared to related work methods (other shapes) on four different tasks (colors).
    The plots represent LPIPS~$\downarrow$ (left), SSIM~$\uparrow$ (center) and PSNR~$\uparrow$ (right) on the y-axis, and processing time~$\downarrow$ (all plots) on the x-axis.
    For better visualization and comparison, each plot is separated into two parts with different scales in the x-axis.
    }
    \label{fig:performance_vs_time_cat}
\end{figure*}

\begin{figure*}[h]
    \centering
    \includegraphics[width=0.99\linewidth]{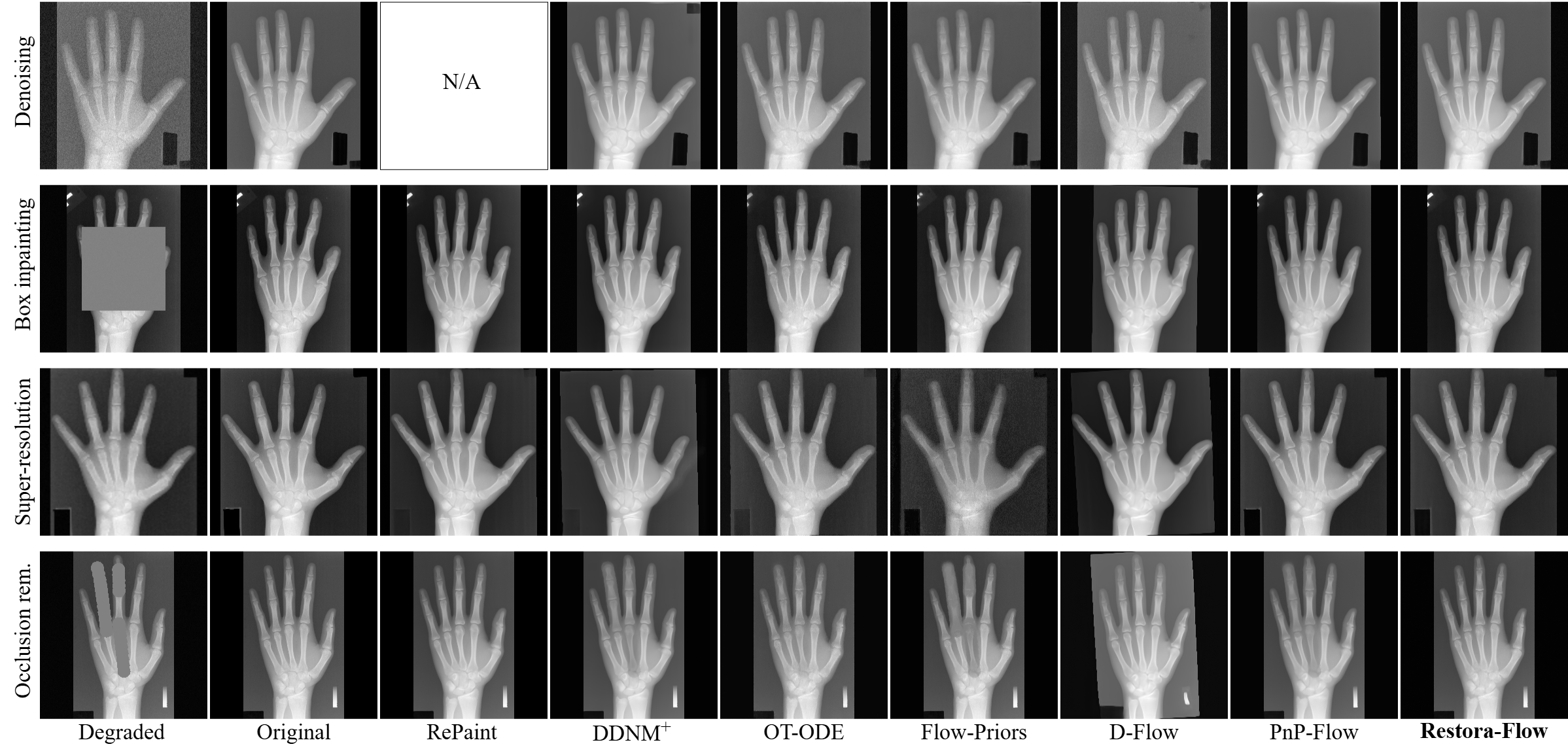}
    \caption{
    Qualitative results on X-ray Hand.
    Shown are the degraded image (col 1), original image (col 2), restored images using related work as indicated (cols 3-8) and restored images of Restora-Flow (col 9).
    Rows refer to different tasks as indicated.  
    RePaint is not applicable to denoising (N/A).
    }
    \label{fig:xray_hand_results}
\end{figure*}

\begin{figure*}[h]
    \centering
    \includegraphics[width=0.98\linewidth]{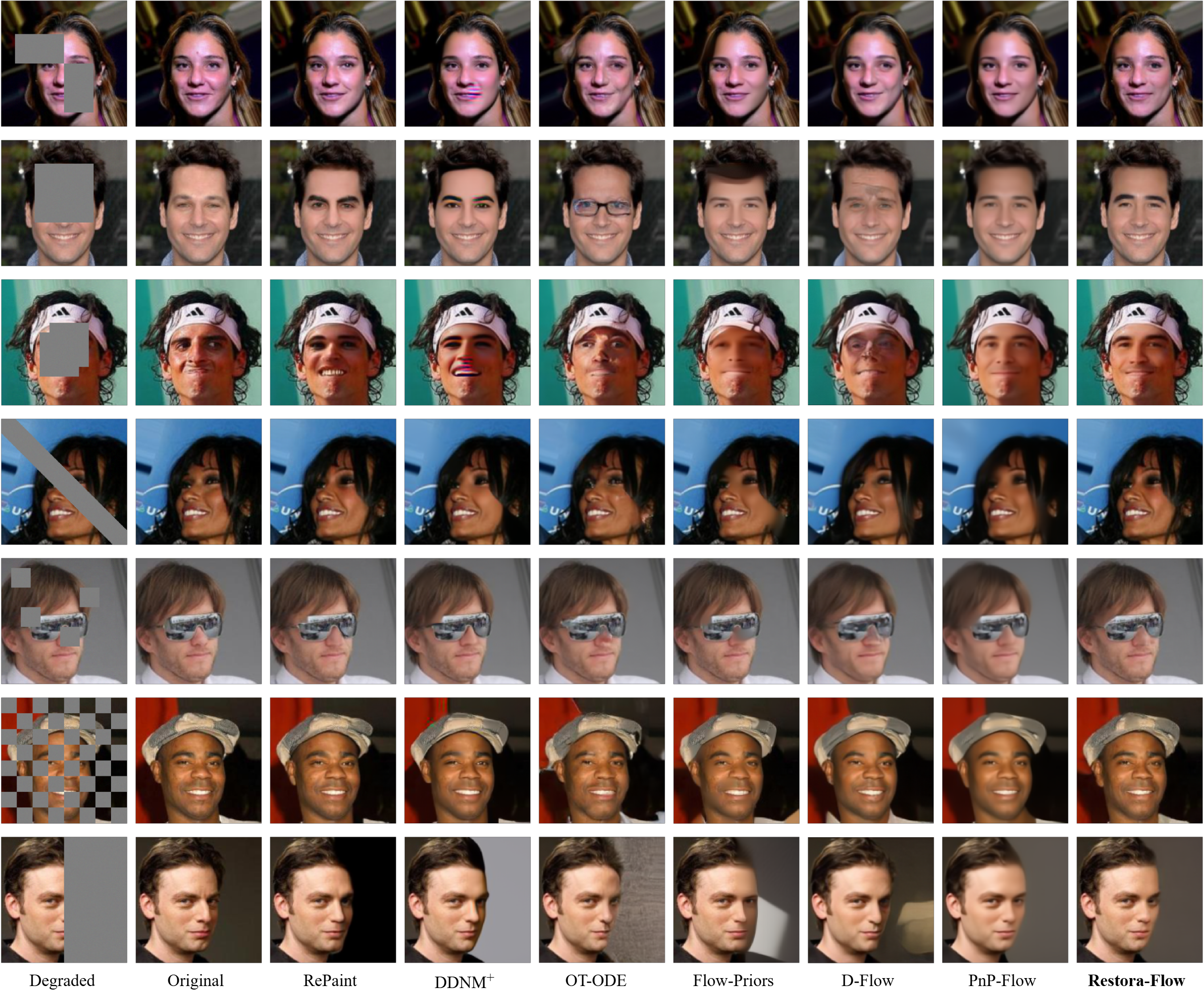}
    \captionof{figure}{
    Qualitative comparison of image restoration methods on CelebA using various input masks. Differences can be best seen in the pdf version.
    }
    \label{fig:celeba_different_masks_suppl}
\end{figure*}

\begin{figure*}[h]
    \centering
    \includegraphics[width=0.98\linewidth]{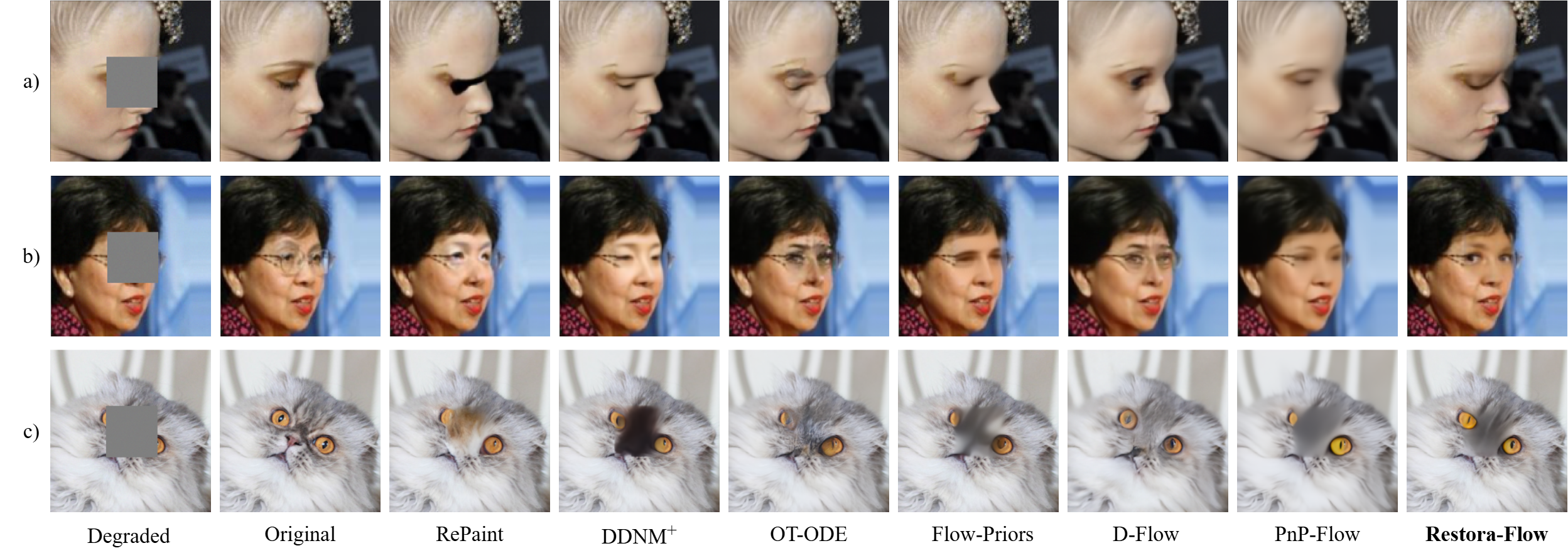}
    \captionof{figure}{
    Representative failure cases of Restora-Flow and comparison methods. (a) A face in profile view, deviating from the typically front-facing images in CelebA. (b) An example with occlusion due to glasses (CelebA). (c) An atypical AFHQ-Cat image where the cat's appearance differs from typical samples in the dataset (\eg, ears not visible).
    }
    \label{fig:failure_cases}
\end{figure*}

\end{document}